\documentclass[sigconf]{acmart}

\AtBeginDocument{%
  \providecommand\BibTeX{{%
    \normalfont B\kern-0.5em{\scshape i\kern-0.25em b}\kern-0.8em\TeX}}}

\setcopyright{acmcopyright}
\copyrightyear{2022}
\acmYear{2022}
\acmDOI{XXXXXXX.XXXXXXX}

\acmConference[ACM Multimedia '2022]{Make sure to enter the correct
  conference title from your rights confirmation emai}{Oct 10--14,
  2022}{lisbon, portugal.}
\acmISBN{978-1-4503-XXXX-X/18/06}

\usepackage{graphicx}
\usepackage{times}
\usepackage{epsfig}
\usepackage{graphicx}
\usepackage{amsmath}
\usepackage{color}
\usepackage{soul}
\usepackage{xcolor}
\usepackage{multirow}
\usepackage{booktabs}
\usepackage{enumerate}
\usepackage{enumitem}
\usepackage{pifont}

\definecolor{mypink}{RGB}{219, 48, 122}
\usepackage[capitalize]{cleveref}
\crefname{section}{Sec.}{Secs.}
\Crefname{section}{Section}{Sections}
\Crefname{table}{Table}{Tables}
\crefname{table}{Tab.}{Tabs.}
\usepackage{comment}

\begin{document}

\title{A Lightweight Graph Transformer Network for Human Mesh Reconstruction from 2D Human Pose}

\author{Ce Zheng}
\affiliation{%
  \institution{University of Central Florida}
  \country{USA}
}
\email{cezheng@knights.ucf.edu}

\author{Matias Mendieta}
\affiliation{%
  \institution{University of Central Florida}
  \country{USA}
}
\email{mendieta@knights.ucf.edu}

\author{Pu Wang}
\affiliation{%
  \institution{ University of North Carolina at Charlotte}
  \country{USA}
}
\email{pu.wang@uncc.edu}

\author{Aidong Lu}
\affiliation{%
  \institution{ University of North Carolina at Charlotte}
  \country{USA}
}
\email{aidong.lu@uncc.edu}

\author{Chen Chen}
\affiliation{%
  \institution{University of Central Florida}
  \country{USA}
}
\email{chen.chen@crcv.ucf.edu}

\renewcommand{\shortauthors}{Ce Zheng et al.}

\begin{abstract}
Existing deep learning-based human mesh reconstruction approaches have a tendency to build larger networks to achieve higher accuracy. Computational complexity and model size are often neglected, despite being key characteristics for practical use of human mesh reconstruction models (e.g. virtual try-on systems). In this paper, we present GTRS, a lightweight pose-based method that can reconstruct human mesh from 2D human pose.
We propose a pose analysis module that uses graph transformers to exploit structured and implicit joint correlations, and a mesh regression module that combines the extracted pose feature with the mesh template to reconstruct the final human mesh.  We demonstrate the efficiency and generalization of GTRS by extensive evaluations on the Human3.6M and 3DPW datasets. In particular, GTRS achieves better accuracy than the SOTA pose-based method Pose2Mesh while only using 10.2\% of the parameters (Params) and 2.5\% of the FLOPs on the challenging in-the-wild 3DPW dataset. \textcolor{magenta}{Code is available at \url{https://github.com/zczcwh/GTRS}} 
\end{abstract}

\maketitle

\section{Introduction}

Analyzing and simulating humans from images is an essential task for computer vision. With the blooming of deep learning methods, human pose estimation (HPE) has been studied extensively and rapid progress has been made. 
In recent years, research in this domain has progressed beyond the estimation of 2D or 3D poses \cite{hrnet}\cite{pavllo20193d}\cite{liu2020attention} with a basic keypoint structure, and the study of reconstructing the entire 3D mesh from a single image has attracted much interest. Mesh representation, which can provide rich human body information and have a better visualization, is more welcomed by real-world applications such as gaming, human-computer interaction, and virtual reality (VR). However, human mesh reconstruction from a single image is a challenging task due to depth ambiguity, occlusion, and complex human body articulation.

Two general approaches exist in the literature for performing mesh reconstruction. One is the direct image-based method, where the pipeline is trained end-to-end from input image to output mesh. The second is to employ an off-the-shelf 2D pose detector as the front end, and design a mesh reconstruction model using 2D poses as input. Most recent progress has been made in the first category, achieving promising performance. However, the performance gain has come at the cost of ever increasing computational requirements and complex models  (for instance, METRO ~\cite{lin2021metro} requires 229M Params and 56.6G FLOPs).
In real-world applications such as human-computer interaction, animated avatar, and VR gaming, the human mesh reconstruction task needs to be efficient and deployable on resource-constrained platforms like VR headsets.

While less studied, pose-based methods are alternative solutions for human mesh recovery with a few advantages for such applications. 
First, pose-based methods provide a modular design that can easily be incorporated with any off-the-shelf 2D pose detectors. With speed as the primary goal, fast pose detectors (e.g. ~\cite{osokin2018real,neff2021efficienthrnet,shen2021towards}) can be deployed on a mobile device in real-time with impressive performance. 
Second, the input to pose-based methods (that is, the detected 2D pose) is extremely sparse data with the size of $J \times 2$, where $J$ is the number of joints. Compared to the image input, it gives more flexibility to design a lightweight mesh reconstruction network to achieve computational and memory efficiency with competitive performance. 
Nonetheless, the existing methods (including the state-of-the-art pose-based method Pose2Mesh ~\cite{Choi_2020_ECCV_Pose2Mesh}) still incur substantial computational and memory overhead. The efficient design of the model is crucial for practical use, but has been almost entirely ignored in the literature.

To bridge this gap, we propose a \textbf{G}raph \textbf{T}ransformer network for human mesh \textbf{R}econ\textbf{S}truction from 2D human pose (GTRS), which is \textbf{the first method} focusing on the efficiency. 
GTRS is a pose-based method designed to fully exploit joint correlations for pose and mesh feature representation while minimizing computational complexity and model size. The operational blocks of GTRS are designed with an intentional combination of graph neural networks and transformer operations.
Recently, graph convolutional networks (GCNs) have shown promising advances in 3D HPE and mesh reconstruction tasks ~\cite{MGCN_2021_ICCV}. 
Human pose data is naturally formulated as a graph, and GCNs can extract useful information with relatively little compute and parameters. Therefore, we harness GCNs to form strong representations from these inherent structural priors on the front-end of GTRS block operations.
After this, these representations are further refined with lightweight transformer structure to powerfully capture global dependencies via its self-attention mechanism.
Thus, these combined operations form our designed graph transformer blocks, which we employ in a parallel fashion for comprehensively exploring human kinematic information from the 2D pose and modeling joint correlations in a lightweight manner.

\begin{figure}[htp]
\vspace{-10pt}
  \centering
  \includegraphics[width=1.0\linewidth]{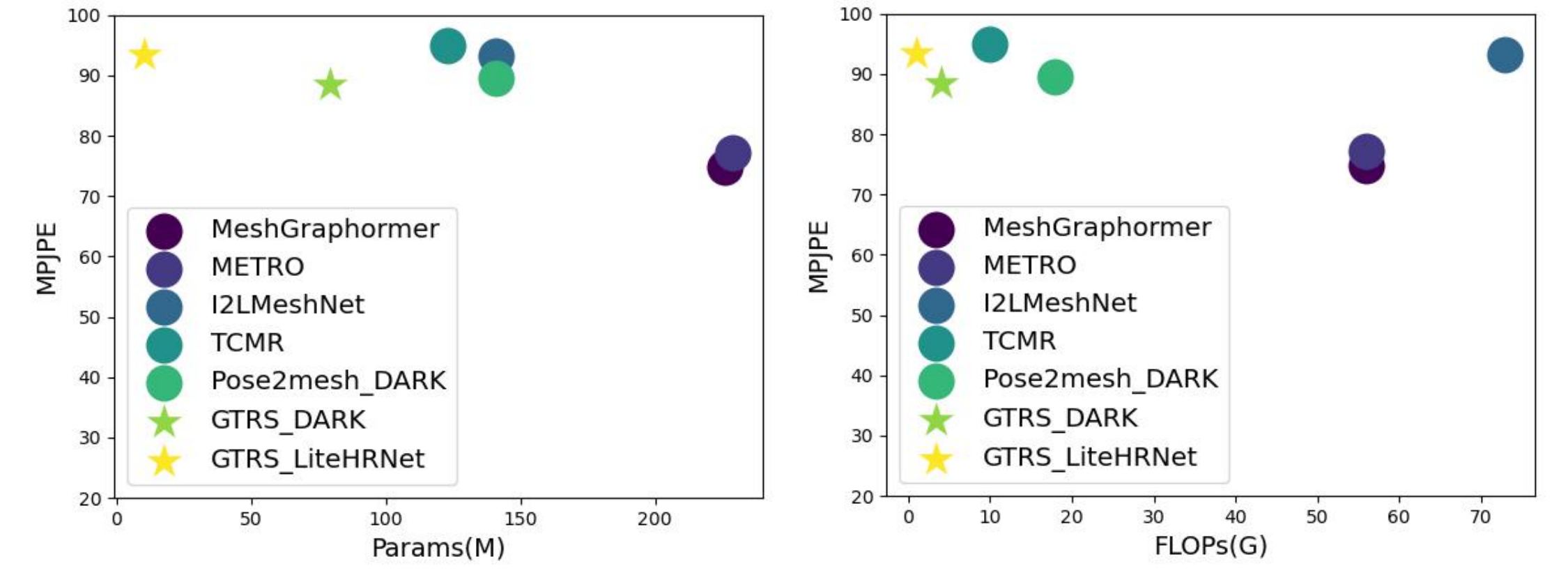}
  \vspace{-15pt}
  \caption{The trade-off between accuracy (MPJPE $\downarrow$) and model Params/FLOPs. All methods are evaluated on 3DPW dataset. GTRS (\textcolor{red}{Ours}) and Pose2Mesh ~\cite{Choi_2020_ECCV_Pose2Mesh} are pose-based methods. The reported Params/FLOPs include the corresponding front-end 2D pose detector (DARK ~\cite{darkpose_2020} or LiteHRNet ~\cite{LiteHRNet}). Others are image-based methods.}
  \label{fig:scatter}
  \vspace{-10pt}
\end{figure}

As an extremely lightweight pose-based method, GTRS uses only 7.9M parameters (Params) and 0.19G floating-point operations (FLOPs) without considering the front-end 2D pose detector. Compared to SOTA pose-based method Pose2Mesh ~\cite{Choi_2020_ECCV_Pose2Mesh}, GTRS achieves better results while only requiring 10.2\% of the Params and 2.5\% of the FLOPs. When also considering the front-end 2D pose detector, GTRS also shows a significant reduction in Params and FLOPs compared to image-based methods in Fig.~\ref{fig:scatter} (e.g. 6.9\% Params and 1.2\% FLOPs compared to I2LMeshNet ~\cite{Moon_I2L_MeshNet}). 
More discussions are provided in Sec. \ref{sec:comapresota}. 

Our contributions are summarized as follows: 
\begin{itemize}[noitemsep,leftmargin=*]  

\item Observing that existing methods mainly pursue higher accuracy while ignoring computational and memory cost, we present a lightweight pose-based method, GTRS, for efficient human mesh reconstruction from the 2D pose. We hope our work can inspire more research on the efficiency of human mesh reconstruction. 

\item We introduce our pose analysis module with a parallel design to facilitate improved utilization of human kinematic information. Within this module, we propose our graph transformer blocks with fixed and learnable adjacency matrices to simultaneously explore diverse structured and implicit human joint correlations.

\item GTRS achieves competitive results compared to previous methods with much fewer parameters and less computational cost on Human3.6M and 3DPW datasets. 
\end{itemize}

\section{Related Work}
\textbf{Human Mesh Reconstruction:}
Recovering human mesh from images is a challenging task that has attracted much attention in recent years. Without requiring additional devices such as depth sensors or inertial measurement
units, Human Mesh Reconstruction (HMR) from images makes it more efficient and convenient. 
The majority of previous works ~\cite{Kolotouros2019SPIN,lin2021metro,lin2021_mesh_graphormer} utilize parametric human model such as SMPL~\cite{SMPL:2015}, ADAM~\cite{TotalCapture}, STAR~\cite{STAR2020} to reconstruct human mesh by training a network to regress model parameters. 

As one of the most popular volumetric models, the SMPL  ~\cite{SMPL:2015} model has been widely used in HMR, e.g., ~\cite{Bogo:ECCV:2016,kanazawaHMR18,Jiang_2020_CVPR,xu2020Low_Resolution}. 
Pavlakos et al. ~\cite{pavlakos2018humanshape}, and Omran et al. ~\cite{omran2018nbf} regress SMPL parameters to reconstruct 3D human mesh. SPIN ~\cite{Kolotouros2019SPIN} revisits optimization approaches within the neural networks that initializes an iterative optimization process (SMPLify). Instead of predicting SMPL parameters, Zhu et al. ~\cite{zhu2019detailed} combine the SMPL model with a hierarchical mesh deformation framework to enhance the flexibility of free-form 3D deformation. Kocabas et al. ~\cite{kocabas2020vibe} include the large-scale motion capture dataset AMASS ~\cite{AMASS2019} for adversarial training of their SMPL-based method named VIBE (Video Inference for Body Pose and Shape Estimation). 

Compared to previous methods that recover human mesh directly from images, ~\cite{Choi_2020_ECCV_Pose2Mesh} estimates SMPL parameters from predicted 2D poses and achieves impressive performance. 
By applying off-the-shelf 2D pose detectors such as AlphaPose ~\cite{fang2017rmpe} and HRNet ~\cite{sun2019deep}, the well-estimated 2D pose can be obtained. Then, a CNN-based PoseNet and MeshNet are proposed to exploit the human mesh topology to recover human mesh based on the input 2D pose. GTRS also follows this pose-based pipeline to reconstruct human mesh. 

\textbf{Graph Convolution Networks:} Recently, graph convolution networks (GCNs) have been widely adopted in 3D human pose estimation (3D HPE) ~\cite{Ci2019,Zhao_2019_Semantic_Graph,Liu_2020_ECCV_weight_sharing,MGCN_2021_ICCV,zhao2021graformer} because of the intuitive modeling of human joints as a graph structure and potential ability to better capture human kinematics.
Following this trend,
GCNs have also gained much attention in human mesh reconstruction ~\cite{kolotouros2019cmr,Choi_2020_ECCV_Pose2Mesh,Graphormer_2021_ICCV}. Kolotouros et al. ~\cite{kolotouros2019cmr} regress the locations of the SMPL mesh vertices using a GCN architecture. Pose2Mesh ~\cite{Choi_2020_ECCV_Pose2Mesh} employs a GCN to regress SMPL parameters from estimated 2D and 3D pose. 
 
\textbf{Vision Transformer:} Transformer architecture is developing rapidly in the field of computer vision. Recent works have demonstrated the powerful global representation ability of transformer attention mechanism in various vision tasks such as object detection ~\cite{carion2020end,zhu2020deformable}, image classification ~\cite{Dosovitskiy2020ViT,liu2021Swin}, segmentation ~\cite{zheng2021rethinking}, human pose estimation ~\cite{Poseformer_2021_ICCV,zhao2021graformer}, 
etc.  Lin et al. ~\cite{lin2021metro} combine CNNs with transformer networks in their method, named METRO, to regress mesh vertices from a single image. 


MeshGraphormer ~\cite{lin2021_mesh_graphormer} is a {close related work}, which also uses GCN with transformer architecture. However, it is \textit{an image-based method} that injects GCN with fixed adjacency matrix into the transformer block between multi-head attention and multilayer perceptron (MLP). \textit{As a pose-based method}, our GTRS utilizes GCNs to model features with prior knowledge, then applies transformers to further explore global dependencies. Moreover, GTRS adopts a paralleled design which enables different graph transformer blocks to explore diverse structured and implicit human kinematic information by using fixed and learnable adjacency metrices.   
Compared to MeshGraphormer ~\cite{lin2021_mesh_graphormer} that requires 226.5M Params and 56.6G FLOPs, GTRS shows significant computational and memory cost reduction. GTRS is more friendly to deploy on mobile devices since it only requires 9.7M Params and 0.89G FLOPs (4.29\% and 1.57\% of MeshGraphormer).


\section{Methodology}

\begin{figure*}[htp]
\vspace{-5pt}
  \centering
  \includegraphics[width=0.8\linewidth]{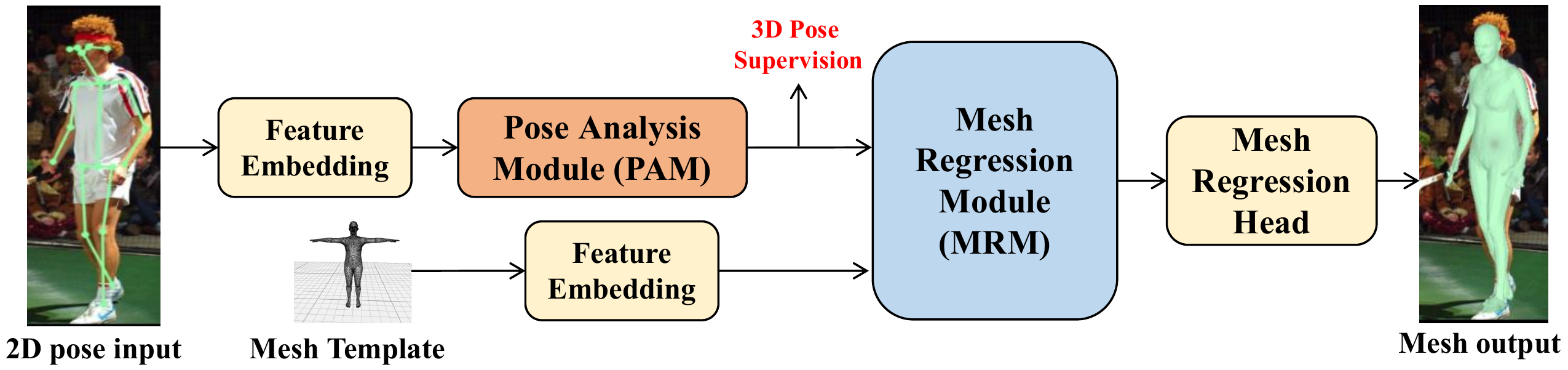}
  \vspace{-8pt}
  \caption{\small{Overview of the proposed GTRS architecture. Given the image, 2D human pose is first detected by an off-the-shelf 2D pose detector. Then, the Pose Analysis Module outputs the pose feature and intermediate 3D pose which is supervised by the ground truth 3D pose.  Next, the pose feature is modeled with template mesh feature in the Mesh Regression Module. Finally, a regression head will output human mesh parameters for reconstruction. The mesh template is provided by ~\cite{Kolotouros2019SPIN} and the mesh template figure is from ~\cite{lin2021metro}. A more detailed illustration of the GTRS architecture is provided in the \textcolor{blue}{appendix}.}}
  \label{fig:architecture}
  \vspace{-5pt}
\end{figure*}

\subsection{Baseline}
\label{sec:Baseline}
In order to achieve model efficiency, an intuitive solution is to utilize the existing lightweight architecture such as MobileNetV2 ~\cite{mobilenetv2} for the 3D human mesh recovery from the 2D pose input. We refer to this simple design as our baseline. However, these CNN-based lightweight architectures are designed to process an image-like input (with the shape of $[C,H,W]$ where $C$ is channel, $H$ is the height, and $W$ is the width). First we embed the 2D pose input of $J$ joints given by $X_{in} \in \mathbb{R} ^{J \times 2} $ to  $X_{baseline} \in \mathbb{R} ^{C \times H \times W} $, then apply MobileNetV2 ~\cite{mobilenetv2} to model mesh features. Finally, the output mesh parameter $Y \in \mathbb{R} ^{6890 \times 3} $ can be estimated after feature embedding.   
However, CNNs do not provide a natural modeling of the graph-like 2D pose input (sparse and meaningfully structured), leaving some representational strength and efficiency on the table.

In light of the limitations of the baseline, we present our proposed GTRS architecture and elaborate the design components in the following.
\subsection{Overview of GTRS}

The vision transformer architecture is designed to capture the global dependencies cross all patches via self-attention mechanism given the input size of $[C, D]$ where $C$ is the number of patches and $D$ is the embedding dimension. Given the 2D pose input  $X_{in} \in \mathbb{R} ^{J \times 2} $, the joint correlations can be exploited when modeling human pose and mesh features through transformer architecture. Therefore, we design a lightweight transformer architecture which is more suitable for modeling 3D human pose and mesh given the 2D pose input rather than a lightweight CNN architecture. The results in Section \ref{sec:comapresota} have proved this claim. 

The overall architecture of GTRS is illustrated in Fig.~\ref{fig:architecture}. The input is the estimated 2D pose obtained by off-the-shelf 2D pose detector such as HRNet ~\cite{hrnet}, which can be denoted as $X_{in} \in \mathbb{R} ^{J \times 2} $, where $J$ is the number of joints. A feature embedding layer embeds input $X_{in} \in \mathbb{R} ^{J \times 2} $ to $X_{pose} \in \mathbb{R} ^{J \times D}$ with a high feature dimension $D$. Then, a Pose Analysis Module (PAM) returns modelled feature $X^{'}_{pose} \in \mathbb{R} ^{J \times D}$. Next, the mesh template  $ M_{temp} \in \mathbb{R} ^{6890 \times 3} $ (from ~\cite{Kolotouros2019SPIN}) is used to provide initial human mesh information, which is embedded to mesh template feature $X_{temp} \in \mathbb{R} ^{T \times D}$ where $T$ is the channel number. 
Then, we feed $X^{'}_{pose}$ and $X_{temp}$ to Mesh Regression Module (MRM), and the output would be $X_{out} \in \mathbb{R} ^{(J+T) \times D} $. Finally, the estimated mesh parameter $Y \in \mathbb{R} ^{6890 \times 3} $ can be obtained after the regression head.

\subsection{Preliminaries of GTRS} 
\textbf{GCN}: 
The GCN was introduced by ~\cite{kipf2017GCN}. A graph is defined as $\mathcal{G} = \{ \nu, \varepsilon \} $ , where $\nu$ is a set of $N$ nodes and $\varepsilon$ is a set of edges. We use GCN to model 2D pose feature, for the input feature $X \in \mathbb{R} ^{J \times D} $, where $J$ is the number of joints and $D$ is the dimension of input feature. Given the adjacency matrix $A \in \mathbb{R} ^{J \times J}$ based on the joints connectivity, the output $X' \in \mathbb{R} ^{J \times D'}$ of one GCN layer can be represented as: 
\begin{align}
\small
    X' = \sigma (AXW)
\end{align}
where $\sigma (\cdot)$ is the activation function for network non-linearity, and $W \in \mathbb{R} ^{D \times D'} $ is the learnable weight matrix which changes the feature dimension from $D$ to $D'$. We use the Gaussian Error Linear Unit (GELU) ~\cite{hendrycks2016gelu} as activation function in this work.  

\noindent\textbf{Transformer}: 
Multi-Head Self-Attention Layer (MHA) is the core function of the transformer blocks, which was proposed by Vaswani et al. ~\cite{vaswani2017attention}. The input $X \in \mathbb{R} ^{J \times D} $ is first mapped to three matrices: query matrix $Q$, key matrix $K$ and value matrix $V$ by three linear transformation: 
\begin{align}
\small
    Q = {X}W_Q, \quad K = {X}W_K, \quad V = {X}W_V.
\end{align}
where $W_Q$, $W_K$ and $W_V$ $\in \mathbb{R} ^{D \times D}$.

The scaled dot product attention can be described as the following mapping function: 
\begin{align}
\small
    {\rm Attention}(Q,K,V) = {\rm Softmax}(QK^\top/ \sqrt{d})V.
\end{align}
where $\frac{1}{\sqrt{d}}$ is the scaling factor for appropriate normalization to prevent extremely small gradients. 

Next, the MHA utilizes multiple heads to model the information jointly from various representation subspaces with different positions. Each head applies scaled dot-product attention in parallel. The MSA output will be the concatenation of $h$ attention head outputs.
\begin{align}
    {{\rm MSA} (Q,K,V)} 
    &= {\rm Concat} (H_1, H_2, \dots, H_h) W_{out} \\  {\rm where} \quad H_i &= {\rm Attention} (Q_i,K_i,V_i), i \in [1,...,h] 
\end{align}
$W_{out}$ is a linear projection $\in \mathbb{R} ^{D \times D}$.  

\subsection{Pose Analysis Module in GTRS}

\begin{figure}[htp]
\vspace{-5pt}
  \centering
  \includegraphics[width=0.9\linewidth]{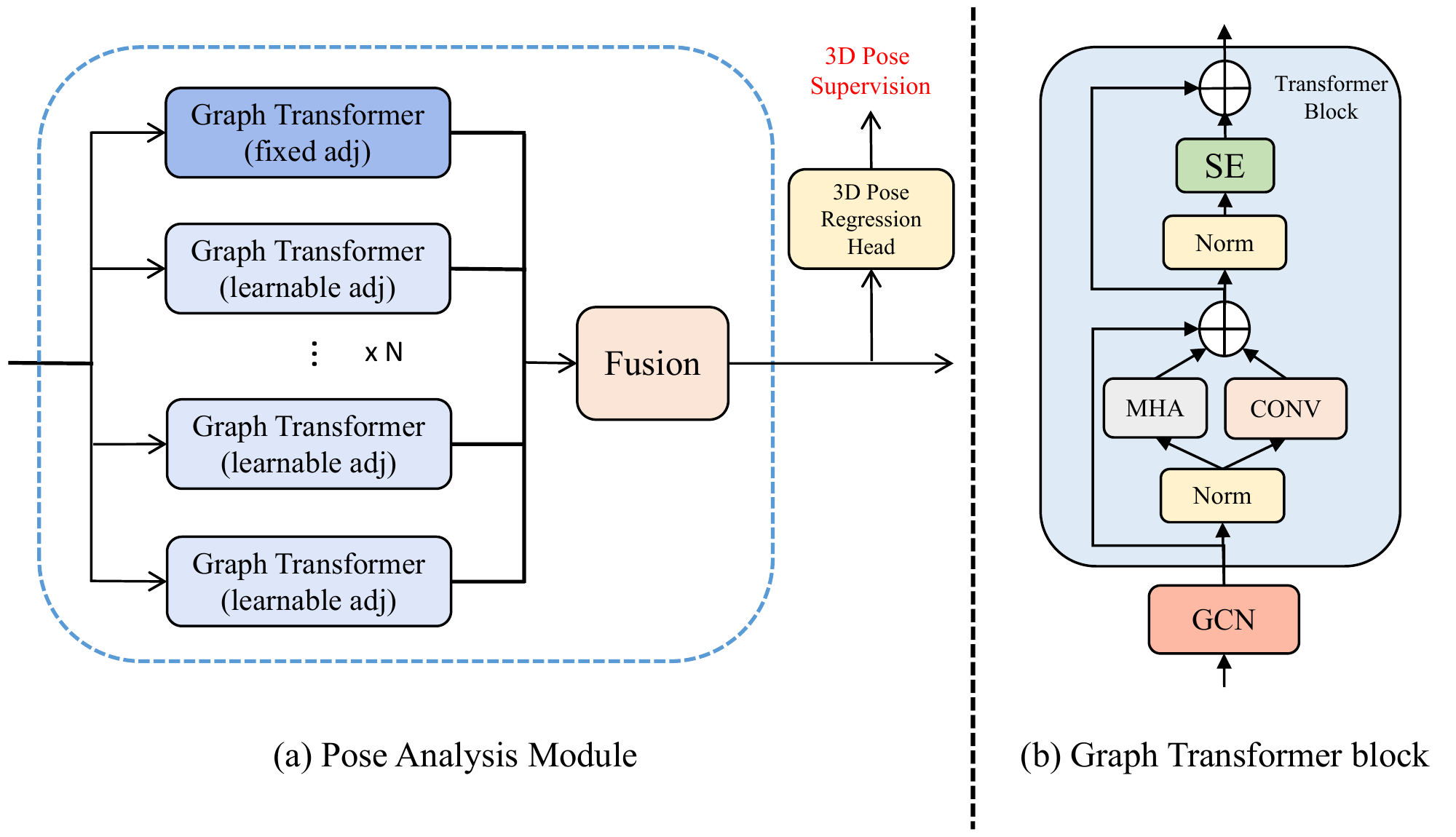}
  \vspace{-10pt}
  \caption{The Pose Analysis Module architecture is shown in (a), which consists of paralleled graph transformer blocks. `fixed adj' means a fixed adjacency matrix is used in this graph transformer block, and `learnable adj' means a learnable adjacency matrix is used. (b) is the architecture of one graph transformer block. }
  \label{fig:PAM}
  \vspace{-5pt}
\end{figure}

\begin{figure}[htp]
  \centering
  \includegraphics[width=0.9\linewidth]{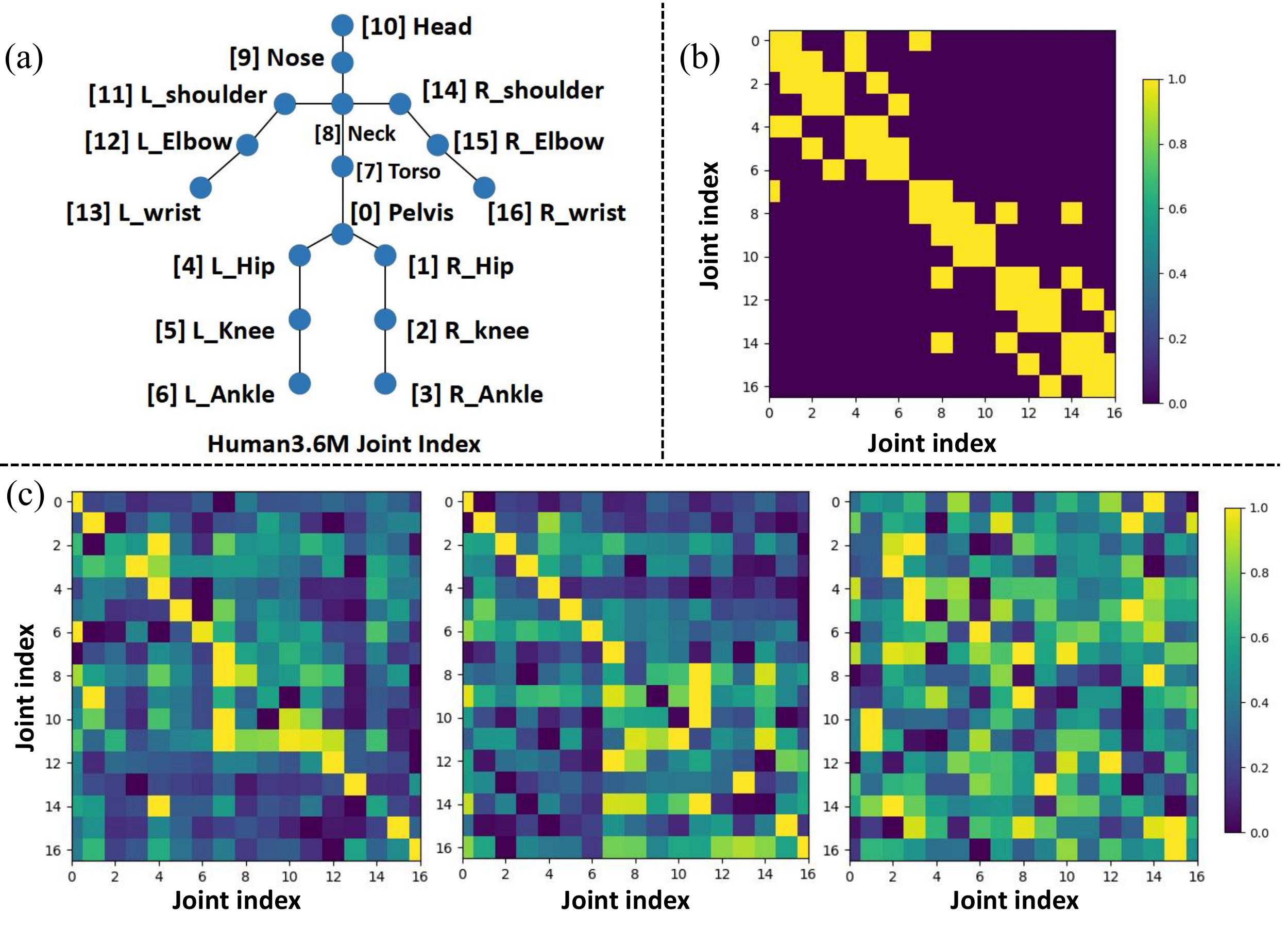}
  \vspace{-10pt}
  \caption{The normalized adjacency matrices used in PAM. (a) is the joint index of Human3.6M dataset. (b) is a fixed adjacency matrix directly from the joint connectivity to model structured correlations. (c) shows learnable adjacency matrices that learned from training data to capture implicit correlations. In PAM, we apply a fixed adjacency matrix and multiple learnable adjacency matrices with a paralleled design to allow the network to explore a diverse set of structured and implicit correlations.} 
  \label{fig:adjvis}
  \vspace{-15pt}
\end{figure}

In the PAM, we utilize graph transformers to improve the structured and implicit correlations based on the human kinematic information. We follow ~\cite{MGCN_2021_ICCV} to build our GCN blocks. 
Then a transformer block is followed to model global dependencies. Our graph transformer block is illustrated in Fig.~\ref{fig:PAM} (b). Different from previous transformer architecture that stacks multiple transformer encoders, we form graph transformer blocks parallelly as shown in Fig.~\ref{fig:PAM} (a). The transformer embedding dimension is set to be a small number to make the network lightweight. 
Because GCNs maintain a strong relationship with the input graph structure, we begin each block with such operations to inject structural priors before performing the transformer's self-attention. 
Based on the human kinematic configuration illustrated in Fig.~\ref{fig:adjvis} (a), the actual adjacency matrix can be obtained as shown in Fig.~\ref{fig:adjvis} (b). Among these paralleled graph transformer blocks, only one block utilizes the actual adjacency matrix, which means this adjacency matrix is fixed and would not be updated. This graph transformer block is maintained to model pose features with structured human kinematic information using the fixed adjacency matrix.
The rest of the graph transformer blocks are responsible to capture implicit correlations from the training data by applying the learnable adjacency matrices. The different patterns of correlations can be discovered through learnable adjacency matrices as shown in Fig.~\ref{fig:adjvis} (c). The parallel design allows for various structural biases to be applied simultaneously. Not only the kinematic correlations can be provided by the fixed adjacency matrix (based on the body joints structure), but also the unexpected correlations beyond our prior knowledge can be discovered by the learnable adjacency matrices in PAM. Next, a fusion block, which is a convolutional layer, will fuse all paralleled features together to a feature that maintains the same size as the input. 

Our transformer encoder layer is different from the original transformer encoder in ~\cite{Dosovitskiy2020ViT}. The structure of our transformer encoder is illustrated in Fig.~\ref{fig:PAM} (b). First, we add one convolutional branch parallel to the MHA branch, which is a point-wise convolution with the $1 \times 1$ convolutional filters as described in ~\cite{howard2017mobilenets}. The reason we use this pointwise convolution is to create linear combinations of the input channels ($J$ joint channels) while maintaining a low computational cost. Then, instead of the MLP layer, we use an SE block ~\cite{hu2018senet} which is also computationally lightweight. The SE block is designed to recalibrate channel-wise feature responses by modeling channel interdependencies. 
The output of our transformer blocks given the input $X_{in} \in \mathbb{R} ^{J \times D} $  can be represented as follows:
\begin{flalign}
\small
   & X' = {\rm MSA}({\rm Norm}(X_{in})) + {\rm CONV}({\rm Norm}(X_{in})) + X_{in} \\
   & X_{out} = {\rm SE}({\rm Norm}(X')) + X'  
\end{flalign}
where  $\rm CONV(\cdot)$  is the convolutional block, $\rm Norm(\cdot)$ is the normalization operator, and $\rm SE(\cdot)$ denotes the SE block. 

Besides the output pose features $X_{out}$, a 3D pose regression head (implemented as an MLP) is used to output a 3D pose prediction $X_{3DPose} \in \mathbb{R} ^{J \times 3}$. This supervision on the 3D pose ensures the PAM can exploit pose features well even if the detected 2D pose is noisy (such as missing joints).



\subsection{Mesh Regression Module in GTRS}
The computational complexity of our transformer structure (as in our PAM) is $\mathcal{O}(nD^{2})$ for the input size of  $X \in \mathbb{R} ^{n \times D}$ and $n$ is much less than $D$. To maintain a lightweight network, the transformer blocks with small embedding dimension $D$ (consistent with the embedding dimension in PAM) are used in MRM. 
Considering that small embedding dimension $D$ may not have enough representation ability, 
we introduce the mesh template which provides rich human mesh information for a better regression. The effectiveness of adding the mesh template has been verified in Sec. \ref{sec:ablation}. We embed the original mesh template to the mesh template feature $X_{temp} \in \mathbb{R} ^{T \times D}$ to reduce the computational cost. 

\begin{figure}[htp]
\vspace{-5pt}
  \centering
  \includegraphics[width=0.9\linewidth]{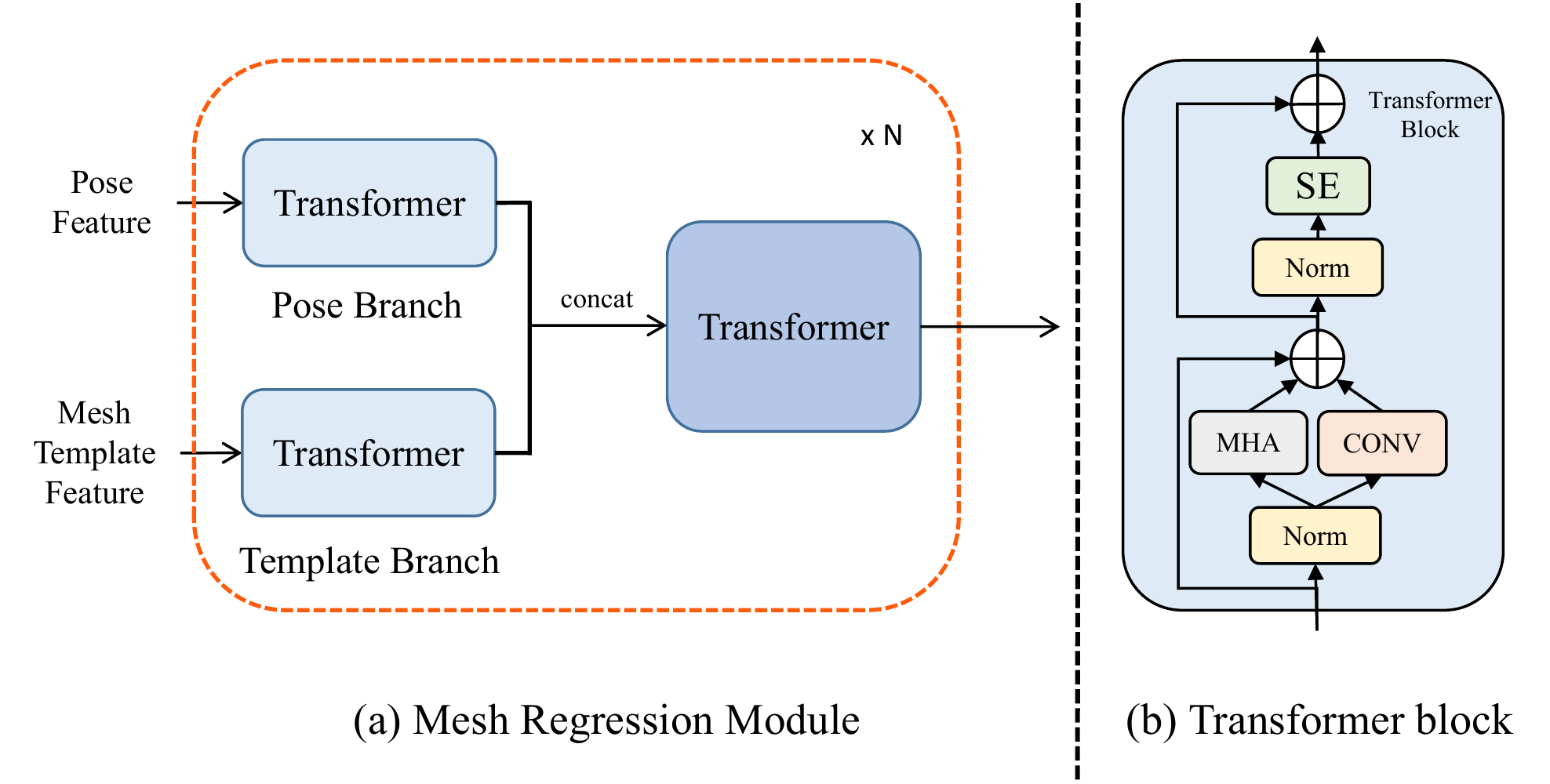}
  \caption{The Mesh Regression Module architecture is shown in (a), which is consists of dual-branch transformer blocks. (b) is the architecture of one transformer block.}
  \label{fig:MRM}
  \vspace{-15pt}
\end{figure}

Due to different scales of the pose feature $X^{'}_{pose} \in \mathbb{R} ^{J \times D}$ and the mesh template feature $X_{temp} \in \mathbb{R} ^{T \times D}$, we design a dual-branch block structure where two separate transformers are applied first to model these two features and then fuse information by a fusion transformer. The output is $X_{fusion} \in \mathbb{R} ^{(J+T) \times D} $, which can be split to $X_{B\_pose} \in \mathbb{R} ^{J \times D}$ and $X_{B\_temp} \in \mathbb{R} ^{T \times D}$ as the input of next block.


Finally, a regression head MLP will upsample the feature $X_{out} \in \mathbb{R} ^{(J+T) \times D}$ to the final mesh output.

\subsection{Loss functions}
For the PAM, we add a linear layer to regress the intermediate 3D human pose (the final 3D human pose is obtained from the final mesh) to get a better pose feature. The pose feature is fed into the MRM. 
To train the PAM, we apply an $L$1 distance loss between the predicted indeterminate 3D pose $J_{3D \_ int} \in  \mathbb{R} ^{K \times 3}$ and the ground truth 3D pose $\hat{J}_{3D} \in  \mathbb{R} ^{K \times 3}$, where $K$ is the number of joints. The indeterminate 3D joint loss is defined as follows:
\begin{align}
\small
    \mathcal{L}_{J \_ int} = \frac{1}{K}\sum_{i=1}^{K}\| J_{3D \_ int} - \hat{J}_{3D} \|_1,
\end{align}

\noindent After pretraining the PAM, we train the entire pipeline includes the PAM and the MRM. The following loss functions are used. 

\textbf{Mesh Vertex Loss}: We use an $L$1 distance loss between the predicted 3D mesh vertices coordinates $V_{3D} \in  \mathbb{R} ^{M \times 3}$ and the ground truth 3D mesh vertices coordinates $\hat{V}_{3D} \in  \mathbb{R} ^{M \times 3}$, where $M$ is the number of vertices. The mesh vertex loss is defined as follows:
\begin{align}
\small
    \mathcal{L}_{Vertex} = \frac{1}{M}\sum_{i=1}^{M}\| V_{3D } - \hat{V}_{3D} \|_1.
\end{align}

\textbf{3D Joint Coordinate Loss}: After estimating 3D mesh vertices coordinates $V_{3D} \in  \mathbb{R} ^{M \times 3}$, we use the model defined joint regression matrix $R \in  \mathbb{R} ^{K \times M}$ to calculate 3D pose based on the estimated mesh, where $K$ is the number of joints and $M$ is the number of vertices. We apply an $L$1 loss between the regressed 3D pose from estimated mesh $RV$ and the ground truth 3D pose $\hat{J}_{3D} \in  \mathbb{R} ^{K \times 3}$. 
\begin{align}
\small
    \mathcal{L}_{Joint} = \frac{1}{K}\sum_{i=1}^{K}\| RV - \hat{J}_{3D} \|_1
\end{align}

\textbf{Surface Normal Loss}: Following ~\cite{wang2018pixel2mesh,Choi_2020_ECCV_Pose2Mesh}, we supervise normal vectors of the estimated mesh surface with the ground truth unit normal vector. We apply this surface normal loss to improve surface smoothness and local detail as in ~\cite{wang2018pixel2mesh}. The surface normal loss is defined as follows:
\begin{align}
\small
    \mathcal{L}_{Normal} = \sum\limits_{f}  \sum\limits_{\{ i,j \} \subset f} \left| \langle \frac{v_i - v_j}{\|v_i - v_j \|_2}, n_{f}^{*} \rangle \right|,
\end{align}
where $f$ denotes a triangle face in the human mesh and $ n_{f}^{*}$ denotes a ground truth unit normal vector of $f$. The $v_i$ and $v_j$ denotes the $i$th and $j$th vertices in $f$, respectively. 

\textbf{Surface Edge Loss}: Following ~\cite{wang2018pixel2mesh,Choi_2020_ECCV_Pose2Mesh}, we use a edge length consistency loss between the predicted edges and ground truth edges. The surface edge loss aims to improve the smoothness of hands, feet, and a mouth which is defined as
\begin{align}
\small
    \mathcal{L}_{Edge} = \sum\limits_{f}  \sum\limits_{\{ i,j \} \subset f} \left|  \|v_i - v_j \|_2 - \| \hat{v_i} - \hat{v_j} \|_2 \right|,
\end{align}
where $f$ denotes a triangle face in the human mesh. The $v_i$ and $v_j$ denotes the $i$th and $j$th vertices in $f$, respectively. \\

\noindent Based on the four type of loss functions, our \textbf{overall loss} is written as
\begin{equation}
\resizebox{.85\hsize}{!}{$
    \mathcal{L} = \lambda_v \mathcal{L}_{Vertex} + \lambda_j \mathcal{L}_{Joint} + \lambda_n \mathcal{L}_{Normal} + \lambda_e \mathcal{L}_{Edge},$}
\end{equation}
where $\lambda_v = 1$, $\lambda_j = 0.01$, $\lambda_n = 0.01$, and $\lambda_e = 0.01$ .   

\section{Experiments}
\subsection{Datasets and Evaluation Metrics}
\textbf{Human3.6M} is one of the most widely used large-scale indoor dataset for 3D HPE and mesh reconstruction ~\cite{h36m_pami}. There are 3.6M video frames recorded by 11 professional actors with performing 17 actions. The ground truth 3D pose annotations were captured by an accurate marker-based motion capture system, but no ground truth 3D mesh annotations were provided. The previous works ~\cite{Kolotouros2019SPIN,kanazawaHMR18,kolotouros2019cmr} used pseudo-ground truth mesh provided by Mosh ~\cite{loper2014mosh}. However, they are no longer accessible due to license issues. Now we use the pseudo-ground truth mesh generated by ~\cite{Choi_2020_ECCV_Pose2Mesh}. Following previous works  ~\cite{Kolotouros2019SPIN,kanazawaHMR18,kolotouros2019cmr,Choi_2020_ECCV_Pose2Mesh}, we select 5 subjects (S1, S5, S6, S7, S8) for training and 2 subjects for testing (S9, S11). 

\textbf{3DPW }is an in-the-wild dataset ~\cite{pw3d2018} that contains 60 video sequences (51K video frames) captured in the outdoor environment. The ground truth 3D pose and mesh annotations are provided. We only use its defined test set for evaluation following ~\cite{Kolotouros2019SPIN,Choi_2020_ECCV_Pose2Mesh}.  

We also use \textbf{MSCOCO}~\cite{lin2014mscoco} and \textbf{MuCo-3DHP}~\cite{muco2018} for mixed training following ~\cite{Choi_2020_ECCV_Pose2Mesh,Kolotouros2019SPIN,Moon_I2L_MeshNet}.
For evaluation metrics, we use the three standard metrics below. The unit for these metrics is millimeters (mm). 

\textbf{MPJPE}: Mean-Per-Joint-Position-Error is used to evaluate the estimated 3D human pose. It is computed as the mean Euclidean distance between the estimated joints and the ground truth joints.

\textbf{PA-MPJPE}: is the MPJPE after Procrustes Analysis ~\cite{grice2009generalized}. The rigid alignment using procrustes analysis is performed the estimated 3D pose, then compute MPJPE with the ground truth 3D pose. P-MPJPE aims to measure the errors of the reconstructed structure without considering translations and rotations.

\textbf{MPVE}: Mean-Per-Vertex-Error is used to evaluate the estimated 3D mesh vertices. It is computed as the mean Euclidean distance between the estimated and the ground truth mesh vertices.

\subsection{Implementation Details}

We implemented GTRS with Pytorch ~\cite{PyTorch} using two NVIDIA RTX 3090 GPUs. GTRS can be trained in an end-to-end manner, but we first pretrain the PAM, then train the entire GTRS after loading those weights for better performance. In the PAM, there are 6 graph transformer blocks (1 with fixed adjacency matrix and 5 with learnable adjacency matrix) and the embedding dimension is 128. We use Adam ~\cite{kingma2014adam} optimizer with a learning rate $1 \times 10^{-4} $ to pretrain the PAM for 120 epochs. In MRM, the embedding dimension is the same as in PAM, which is 128, and the layer number $N$ is 4. The mesh template $ M_{temp} \in \mathbb{R} ^{6890 \times 3} $ is the mean SMPL parameters provided by ~\cite{Kolotouros2019SPIN}. We also use Adam ~\cite{kingma2014adam} optimizer with a learning rate $1 \times 10^{-4} $ to train the GTRS for 180 epochs. The total training time would be less than one day. 
\subsection{Comparison with state-of-the-art results}
\label{sec:comapresota}

\begin{table}[]
\scriptsize
\vspace{-5pt}
\centering
  \caption{  Performance comparison with SOTA methods on Human3.6M. ``$\dagger$'' denotes using GT (ground truth) 2D pose as input.}
  \vspace{-5pt}
  \resizebox{0.98\linewidth}{!}{
  \begin{tabular}{c|cc|cc}
\hline
                                                                       & \multicolumn{2}{c|}{\multirow{2}{*}{Methods}}   & \multicolumn{2}{c}{Human3.6M} \\ \cline{4-5} 
                                                                       & \multicolumn{2}{c|}{}                           & MPJPE$\downarrow$         & PA-MPJPE$\downarrow$     \\ \hline
\multirow{6}{*}{\begin{tabular}[c]{@{}c@{}}Image\\ Based\end{tabular}} & \multicolumn{1}{l|}{HMR~\cite{kanazawaHMR18}}            & CVPR 2018 & 88.0          & 56.8          \\
                                                                       & \multicolumn{1}{l|}{GraphCMR~\cite{kolotouros2019cmr}}       & CVPR 2019 & -             & 50.1          \\
                                                                       & \multicolumn{1}{l|}{SPIN~\cite{Kolotouros2019SPIN}}           & ICCV 2019 & -             & 41.1          \\
                                                                       & \multicolumn{1}{l|}{METRO~\cite{lin2021metro}}          & CVPR 2021 & 54.0          & 36.7          \\
                                                                       & \multicolumn{1}{l|}{MeshGraphormer~\cite{lin2021_mesh_graphormer}} & ICCV 2021 & \textbf{51.2} & \textbf{34.5} \\
                                                                       & \multicolumn{1}{l|}{PyMAF~\cite{pymaf2021}}          & ICCV 2021 & 57.7          & 40.5          \\ \hline
\multirow{2}{*}{\begin{tabular}[c]{@{}c@{}}Video\\ Based\end{tabular}} & \multicolumn{1}{l|}{VIBE~\cite{kocabas2020vibe}}           & CVPR 2020 & 65.6          & 41.4          \\
                                                                       & \multicolumn{1}{l|}{TCMR~\cite{TCMR_Choi_2021}}           & CVPR 2021 & 62.3          & 41.1          \\ \hline
\multirow{3}{*}{\begin{tabular}[c]{@{}c@{}}Pose\\ Based \\ (detected by ~\cite{sun2019deep})\end{tabular}}  & \multicolumn{1}{l|}{Pose2Mesh~\cite{Choi_2020_ECCV_Pose2Mesh}}      & ECCV 2020 & 64.9          & 47.0          \\
                                                                       & \multicolumn{1}{l|}{Baseline }           &   -        & 68.3          & 50.0          \\
                                                                       & \multicolumn{1}{l|}{GTRS}           &   -        & 64.3          & 45.4          \\
                                                                       \hline

\hline

\hline
\multirow{3}{*}{\begin{tabular}[c]{@{}c@{}}Pose\\ Based \\(GT 2D)\end{tabular}} & \multicolumn{1}{l|}{Pose2Mesh~\cite{Choi_2020_ECCV_Pose2Mesh}$\dagger$}      & ECCV 2020 & 51.3          & 34.9          \\
                                                                       & \multicolumn{1}{l|}{Baseline$\dagger$ }           &   -        & 57.5          & 39.6         \\
                                                                       & \multicolumn{1}{l|}{GTRS$\dagger$}           &   -        & \textbf{50.3}           & \textbf{30.4}           \\
                                                    \hline
\end{tabular}
}
\label{tab: h36mresults}
\vspace{-10pt}
\end{table}
\textbf{Evaluation on Human3.6M}:
Table \ref{tab: h36mresults} compares GTRS with previous SOTA image/video-based and pose-based methods on Human3.6M test set. 
We follow the same setting as Pose2Mesh ~\cite{Choi_2020_ECCV_Pose2Mesh} and ~\cite{Kolotouros2019SPIN,kanazawaHMR18,kolotouros2019cmr} for a fair comparison; that is, we only use the Human3.6M training set for training and PA-MPJPE is only measured on the frontal camera set. Same as Pose2Mesh ~\cite{Choi_2020_ECCV_Pose2Mesh}, ~\cite{sun2019deep} is used as the 2D pose detector. Within the similar total number of parameters and FLOPs, the performance boost from our baseline to GTRS demonstrates that transformer architecture is more suitable than CNN-based architecture when considering model efficiency for pose-based human mesh recovery. GTRS achieves better performances than Pose2Mesh ~\cite{Choi_2020_ECCV_Pose2Mesh} (MPJPE decreases 0.6 and PA-MPJPE decreases 1.6). Compared with video-based methods, GTRS still achieves similar MPJPE (within 2 points) as a lightweight model. 
When using ground truth 2D pose as input, GTRS outperforms previous methods both in terms of MPJPE and PA-MPJPE. This results indicates the lower bound of GTRS. As more accurate and robust 2D human pose detectors are proposed, they can be plugged-in to GTRS and close the gap towards this lower bound. 


\begin{table}[]
\scriptsize
\centering
\vspace{-5pt}
  \caption{Performance comparison with SOTA methods on 3DPW dataset. * indicates 3DPW training set is used during training.  ``$\dagger$'' denotes using GT 2D pose as input.
  \vspace{-5pt}
}
  \resizebox{\linewidth}{!}{
  \begin{tabular}{c|cc|ccc}
\hline
\multirow{2}{*}{}                                                               & \multicolumn{2}{c|}{\multirow{2}{*}{Methods}}   & \multicolumn{3}{c}{3DPW dataset}              \\ \cline{4-6} 
                                                                                & \multicolumn{2}{c|}{}                           & MPJPE$\downarrow$         & PA-MPJPE$\downarrow$      & MPVE$\downarrow$         \\ \hline
\multirow{6}{*}{\begin{tabular}[c]{@{}c@{}}Image\\ Based\end{tabular}}          & \multicolumn{1}{c|}{HMR~\cite{kanazawaHMR18}}            & CVPR 2018 & -             & 81.3          & -             \\
                                                                                & \multicolumn{1}{c|}{GraphCMR~\cite{kolotouros2019cmr}}       & CVPR 2019 & -             & 70.2          & -             \\
                                                                                & \multicolumn{1}{c|}{SPIN~\cite{Kolotouros2019SPIN}}           & ICCV 2019 & -             & 59.2          & 116.4         \\
                                                                                & \multicolumn{1}{c|}{I2LMeshNet~\cite{Moon_I2L_MeshNet} }        & ECCV 2020 & 93.2          & 57.7          & -             \\
                                                                                & \multicolumn{1}{c|}{PyMAF~\cite{pymaf2021}}          & ICCV 2021 & 92.8          & 58.9          & 110.1         \\
                                                                                & \multicolumn{1}{c|}{MeshGraphormer*~\cite{lin2021_mesh_graphormer} } & ICCV 2021 & \textbf{74.7} & \textbf{45.6} & \textbf{87.7} \\ \hline
\multirow{3}{*}{\begin{tabular}[c]{@{}c@{}}Video\\ Based\end{tabular}}          & \multicolumn{1}{c|}{VIBE~\cite{kocabas2020vibe}}           & CVPR 2020 & 93.5         & 56.5          & 113.4          \\
                                                                                & \multicolumn{1}{c|}{VIBE*~\cite{kocabas2020vibe}}           & CVPR 2021 & 82.9          & 51.9          & 99.1         \\
                                                                                & \multicolumn{1}{c|}{TCMR~\cite{TCMR_Choi_2021}}           & CVPR 2021 & 95.0          & 55.8          & 111.5         \\ \hline
\multirow{3}{*}{\begin{tabular}[c]{@{}c@{}}Pose\\ Based\\(detected by ~\cite{darkpose_2020})\end{tabular}}           & \multicolumn{1}{c|}{Pose2Mesh~\cite{Choi_2020_ECCV_Pose2Mesh}}      & ECCV 2020 & 88.9          & 58.3          & 106.3         \\
                                                                                & \multicolumn{1}{c|}{Baseline}       &     -      &   95.6            &   61.3            &    129.8           \\
                                                                                & \multicolumn{1}{c|}{GTRS}           &    -       & 88.5          & 58.9          & 106.2         \\ 
                                                                       \hline

\hline

\hline
\multirow{3}{*}{\begin{tabular}[c]{@{}c@{}}Pose\\ Based\\ (GT 2D)\end{tabular}} & \multicolumn{1}{c|}{Pose2Mesh~\cite{Choi_2020_ECCV_Pose2Mesh}$\dagger$ }      & ECCV 2020 & 65.1          & 34.6          & -             \\
                                                                                & \multicolumn{1}{c|}{Baseline$\dagger$ }       &   -        & 67.7          & 36.1          & 70.3          \\
                                                                                & \multicolumn{1}{c|}{GTRS$\dagger$ }           &     -      & \textbf{53.8} & \textbf{34.5} & \textbf{61.6} \\ \hline
\end{tabular}
}
\label{tab: 3dpwresults}
\vspace{-15pt}
\end{table}

\textbf{Evaluation on 3DPW}:
Table \ref{tab: 3dpwresults} compares GTRS with previous SOTA image/video-based and pose-based methods on 3DPW test set. 
The training sets include Human3.6M, COCO, MPII ~\cite{mpii}, UP3D ~\cite{UP3D}, MPI-INF-3DHP ~\cite{MPIINF}, and AMASS~\cite{AMASS2019}. Each method uses a different combination of these datasets. For GTRS, we only use Human3.6M  ~\cite{h36m_pami}, COCO  ~\cite{lin2014mscoco}, and MuCo ~\cite{muco2018} as the training sets. Following Pose2Mesh ~\cite{Choi_2020_ECCV_Pose2Mesh}, we use DARK ~\cite{darkpose_2020} as the 2D pose detector for 3DPW evaluation. Comparing our baseline with GTRS under similar total number of parameters and FLOPs, we can draw the same conclusion that transformer architecture is more suitable than CNN-based architecture when considering model efficiency for pose-based human mesh recovery.
Note here METRO ~\cite{lin2021metro} and MeshGraphormer ~\cite{lin2021_mesh_graphormer} used 3DPW training set while others did not. Among those methods without using 3DPW training set, GTRS achieves comparable results with much less computational cost.

When using ground truth 2D pose input, the performance of GTRS can be improved significantly. GTRS yields the lowest MPJPE of 53.8, PA-MPJPE of 34.5, and MPVPE of 61.6 (more than 24\% reduction compared with MeshGraphormer ~\cite{lin2021_mesh_graphormer}). 3DPW dataset is a challenging in-the-wild dataset, distinct from the lab-controlled Human3.6M dataset. Occlusions and atypical human postures in these in-the-wild cases are the biggest challenges for accurate 3D human mesh recovery. Due to the difficulties of obtaining sufficient training data with accurate mesh annotations, these issues are difficult to address directly in 3D pose estimation.
However, it is much easier to acquire sufficient training data with accurate 2D pose annotations. 
By simply plugging-in a robust and more accurate 2D pose detector in the future, GTRS can further improve the performance and approach the lower bound.  Therefore, GTRS has the potential to continue staying relevant as an effective approach for 3D mesh reconstruction as 2D pose estimators inevitably improve.


\begin{table*}[]
\scriptsize
\centering
\vspace{-5pt}
  \caption{ The Params and FLOPs comparison on 3DPW dataset. * indicates 3DPW training set is used during training.
  \vspace{-5pt}
}
  \resizebox{\linewidth}{!}{
  \begin{tabular}{c|cc|cc|cc|cc|c}
\hline
                                                                       & \multicolumn{2}{c|}{\multirow{2}{*}{Methods}}             & \multicolumn{2}{c|}{Feature extractor} & \multicolumn{2}{c|}{Proposed model} & \multicolumn{2}{c|}{Overall} & MPJPE \\
                                                                       & \multicolumn{2}{c|}{}                                     & Params (M)             & FLOPs (G)            & Params (M)           & FLOPs (G)           & Params (M)        & FLOPs (G)       & on 3DPW  \\ \hline
\multirow{3}{*}{\begin{tabular}[c]{@{}c@{}}Image\\ Based\end{tabular}} & \multicolumn{1}{l|}{I2LMeshNet~\cite{Moon_I2L_MeshNet}}                  & ECCV2020  & -                  & -                 & -                & -                & 140.5        & 73.2        & 93.2     \\
                                                                       & \multicolumn{1}{l|}{METRO*~\cite{lin2021metro}}                    & CVPR2021  & -                  & -                 & -                & -                & 229.2          & 56.6       & 77.1     \\
                                                                       & \multicolumn{1}{l|}{MeshGraphormer*~\cite{lin2021_mesh_graphormer}}           & ICCV2021  & -                  & -                 & -                & -                & 226.5        & 56.6        & \textbf{74.7}      \\ \hline
\multirow{2}{*}{\begin{tabular}[c]{@{}c@{}}Video\\ Based\end{tabular}} & \multicolumn{1}{l|}{VIBE~\cite{kocabas2020vibe}}                     & CVPR2020 & 25.6              & 8.2             & 33.0            & 1.3             & 58.6         & 9.6         & 93.5     \\
                                                                       & \multicolumn{1}{l|}{TCMR~\cite{TCMR_Choi_2021}}                     & CVPR2021  & 25.6              & 8.2             & 97.3            & 2.1             & 122.9          & 10.3        & 95.0     \\ \hline
                                                                       & \multicolumn{2}{c|}{\multirow{2}{*}{Methods}}             & \multicolumn{2}{c|}{2D pose detector}  & \multicolumn{2}{c|}{Proposed model} & \multicolumn{2}{c|}{Overall} & MPJPE \\
                                                                       & \multicolumn{2}{c|}{}                                     & Param (M)              & FLOPs (G)             & Param (M)            & FLOPs (G)           & Param (M)        & FLOPs (G)       & on 3DPW  \\ \hline
\multirow{6}{*}{\begin{tabular}[c]{@{}c@{}}Pose\\ Based\end{tabular}}  & \multicolumn{1}{l|}{Pose2Mesh with DARK~\cite{darkpose_2020}}  & ECCV2020  & 63.6              & 3.6              & 77.1              & 7.5              & 140.7        & 11.1        & 88.9     \\
                                                                       & \multicolumn{1}{l|}{Baseline with DARK~\cite{darkpose_2020}}        &   -        & 63.6              & 3.6              &8.3             & 0.32             & 71.9         & 3.9         & 97.7     \\
                                                                       & \multicolumn{1}{l|}{GTRS with DARK~\cite{darkpose_2020}}        &   -        & 63.6              & 3.6              &\textbf{7.9}             & \textbf{0.19}             & 71.5         & 3.8         & 88.5     \\ \cline{2-10} 
                                                                       & \multicolumn{1}{l|}{Pose2Mesh with LiteHRNet~\cite{LiteHRNet}} & ECCV2020  & \textbf{1.8}               & \textbf{0.7}               & 77.1              & 7.5              & 78.9         & 8.2        & -        \\
                                                                       & \multicolumn{1}{l|}{Baseline with LiteHRNet~\cite{LiteHRNet}}        &   -        & \textbf{1.8}               & \textbf{0.7}                          &8.3             & 0.32             & 11.1         & 1.1         & 103.4     \\
                                                                       & \multicolumn{1}{l|}{GTRS with LiteHRNet~\cite{LiteHRNet}}       &    -       & \textbf{1.8}               & \textbf{0.7}               & \textbf{7.9}             & \textbf{0.19}            & \textbf{9.7}           & \textbf{0.89}       & 93.5     \\ \hline
\end{tabular}
}
\label{tab: flopsresults}
\vspace{-5pt}
\end{table*}
\textbf{Model Size and FLOPs Comparison:} \label{sec:FLOPs}
Previous human mesh reconstruction methods did not pay much attention to model efficiency. These methods mainly pursued higher accuracy without considering computation and memory. Table \ref{tab: flopsresults} reports the Params and FLOPs comparison between previous methods and GTRS. Pose-based methods can easily select a lightweight 2D pose detector to reduce the computational burden. But for most image/video-based methods, a very large feature extractor is needed for extracting features (e.g. ResNet-50 is used by ~\cite{TCMR_Choi_2021,kocabas2020vibe,Kolotouros2019SPIN}). GTRS only requires 7.9M Params (10.2\% of the Pose2Mesh ~\cite{Choi_2020_ECCV_Pose2Mesh}) and 0.19G FLOPs (2.5\% of the Pose2Mesh ~\cite{Choi_2020_ECCV_Pose2Mesh}) while achieving better results than Pose2Mesh ~\cite{Choi_2020_ECCV_Pose2Mesh} when using the same 2D pose input detected by DARK ~\cite{darkpose_2020}. When compared with our baseline, GTRS improves the performance while maintaining lower memory and computational cost. When we use Lite-HRNet ~\cite{LiteHRNet} as 2D pose detector, the overall Params is 9.7M and FLOPs is 0.89G. GTRS achieves close results with significant Params and FLOPs reduction (6.9\% and 1.2\% of I2LMeshNet ~\cite{Moon_I2L_MeshNet}, 16.5\% and 9.2\% of VIBE  ~\cite{kocabas2020vibe}, 7.9\% and 8.6\% of TCMR ~\cite{TCMR_Choi_2021}). It also can be observed that METRO ~\cite{lin2021metro} and MeshGraphormer ~\cite{lin2021metro} demanded  an extremely large number of Params and FLOPs to achieve the SOTA results; they also used the 3DPW training set while other methods did not. Apart from that, METRO ~\cite{lin2021metro} and MeshGraphormer ~\cite{lin2021metro} were trained in 5 days on 8 NVIDIA V100 GPUs, which is incredibly time and resource consuming. GTRS, on the other hand, can be trained in less than one day on two NVIDIA RTX 3090 GPUs. \textbf{To summarize, GTRS is much more time and resource efficient compared to image-based methods and SOTA pose-based method.} 

\subsection{Ablation Study} \label{sec:ablation}
We conduct the ablation study on  \textcolor{blue}{Human3.6M dataset} (training on S1, S5, S6, S7, S8, and testing on S9 and S11) and report the accuracy using MPJPE and PA-MPJPE.

\begin{table}[]
\tiny
\centering
\vspace{-5pt}
  \caption{ Ablation study on different components in PAM.
  \vspace{-5pt}
}
  \resizebox{1\linewidth}{!}{
  \begin{tabular}{cccc}
\hline
\multicolumn{4}{c}{Architecture in PAM}                                                                                                                                                                       \\ \hline
\begin{tabular}[c]{@{}c@{}}graph transformer blocks\\  with fixed adj\end{tabular} & \multicolumn{1}{c|}{\begin{tabular}[c]{@{}c@{}}graph transformer blocks \\ with learnable adj\end{tabular}} & MPJPE$\downarrow$          & PA-MPJPE$\downarrow$       \\ \hline
1                                                                        & \multicolumn{1}{c|}{0}                                                                             & 68.0          & 50.3          \\
1                                                                        & \multicolumn{1}{c|}{1}                                                                             & 66.9          & 48.9          \\
1                                                                        & \multicolumn{1}{c|}{3}                                                                             & 65.1          & 48.2          \\
1                                                                        & \multicolumn{1}{c|}{5}                                                                             & \textbf{64.3} & \textbf{47.5} \\
1                                                                        & \multicolumn{1}{c|}{7}                                                                             & 64.6          & 47.6          \\
6                                                                        & \multicolumn{1}{c|}{0}                                                                             & 65.3          & 48.4          \\
0                                                                        & \multicolumn{1}{c|}{6}                                                                             & 66.4          & 48.8          \\ \hline
\multicolumn{2}{c|}{pure transformer blocks}                                                                                                                                  &           &      \\ \hline
\multicolumn{2}{c|}{6}                                                                                                                                                        & 67.2          & 49.5          \\ \hline
\end{tabular}
}
\label{tab: Ab_GCN}
\vspace{-10pt}
\end{table}

\textbf{Effectiveness of Using Graph Transformer Blocks:}
We investigate the use of graph transformer blocks in the PAM in Table \ref{tab: Ab_GCN}. All blocks are applied in parallel as shown in Fig ~\ref{fig:PAM}. When we only use one fixed graph transformer block, the MPJPE is 68.0 and PA-MPJPE is 50.3. 
By adding learnable graph transformer blocks, and thereby enabling the exploration of implicit correlations, the performance improves.
The network achieves the best results (MPJPE is 64.3 and PA-MPJPE is 47.5) when using one graph transformer block with a fixed adjacency matrix and five graph transformer blocks with learnable adjacency matrices, 

Next, to verify that the observed performance increase is not solely a matter of additional blocks, we also investigate fixing all adjacent matrices in the six graph transformer blocks. This enforces the graph transformer blocks only to learn with structured correlations of human kinematics. The MPJPE is 65.3 and PA-MPJPE is 48.4. On the contrary, we then try setting all adjacent matrices in the six graph transformer blocks to be learnable during the training.
This allows the graph transformer blocks to learn implicit correlations of human kinematics, and results in a MPJPE of 66.4 and PA-MPJPE of 48.8. Lastly, we apply six pure transformer blocks (without any GCNs), which gives a MPJPE of 67.2 and PA-MPJPE of 49.5. All these results are worse than applying six graph transformer blocks (one with fixed adjacency matrix and five with learnable adjacency matrices), which verifies the effectiveness of using graph transformer blocks with both fixed and learnable adjacency matrices.



\textbf{Impact of 2D Pose Detectors:}
In Table \ref{tab: Ab_noise}, we analyze the impact of the quality of 2D pose on the final mesh performance. 
When using ground-truth 2D pose as input, GTRS can get 50.3 of MPJPE and 30.4 of PA-MPJPE, which outperform the Pose2Mesh ~\cite{Choi_2020_ECCV_Pose2Mesh} results (51.3 of MPJPE and 34.9 of PA-MPJPE). Here, we see that accuracy of GTRS can be improved when using more precise 2D pose inputs.
We also evaluate the performance of using different 2D pose detectors. Incorporated with ~\cite{get2dpose}, the MPJPE of GTRS is 64.3 and PA-MPJPE is 47.5, but the Params is 34M and FLOPs is 6.4G. When switching to a more lightweight 2D pose detector  ~\cite{LiteHRNet}, the entire pipeline is more computational and memory efficient (Params is 9.7M and FLOPs is 0.9G) while preserving the accuracy.


\textbf{Impact of 2D Pose Input during Inference:}
For testing in-the-wild images, the quality of the 2D pose would be affected by various fluctuations based on the input image. We evaluate the robustness of our trained model against potential perturbations since GTRS relies on the input pose. 

\begin{figure}[htp]
\vspace{-5pt}
  \centering
  \includegraphics[width=1\linewidth]{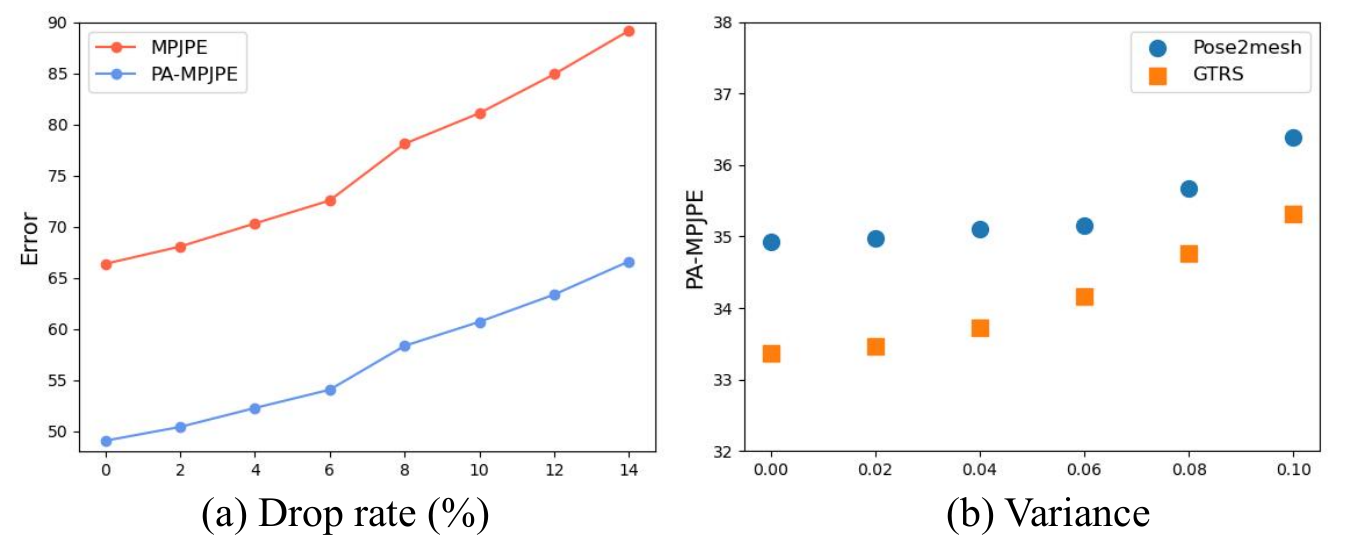}
  \vspace{-15pt}
  \caption{(a) Impact of missing joints during inference. Each joint has a drop probability to simulate missing joints by the 2D pose detector during inference.  (b) Impact of noisy 2D pose input during inference. Various degrees of Gaussian noise (i.e. variance $\sigma ^2$) are added to the input 2d pose to simulate in-the-wild 2D pose input during inference. $\sigma ^2=0$ means GT pose provided in the Human3.6M dataset.}
  \vspace{-5pt}
  \label{fig:g_noise}
  \vspace{-5pt}
\end{figure}

First, we evaluate if GTRS can still perform well when input joints are missing. During inference, we set a drop probability for the joints and the results are shown in Fig.~\ref{fig:g_noise} (a). In PAM, we output the intermediate 3D pose for a 3D pose supervision, which enables GTRS can still achieve acceptable results given a large drop rate.  

Second, we evaluate the robustness of GTRS to noisy 2D pose input. Specifically, we add Gaussian noise $\mathcal{N} (0, \sigma ^2)$ to the input 2d pose to simulate in-the-wild 2D pose input during inference. We do not retrain the model (which is trained using GT pose), instead, we directly evaluate the performance of noisy pose input on the trained model. The results are shown in Fig.~\ref{fig:g_noise} (b). GTRS consistently outperforms Pose2Mesh ~\cite{Choi_2020_ECCV_Pose2Mesh}, demonstrating that GTRS is more robust to in-the-wild inference.  

\begin{figure*}[htp]
\vspace{-5pt}
  \centering
  \includegraphics[width=0.91\linewidth]{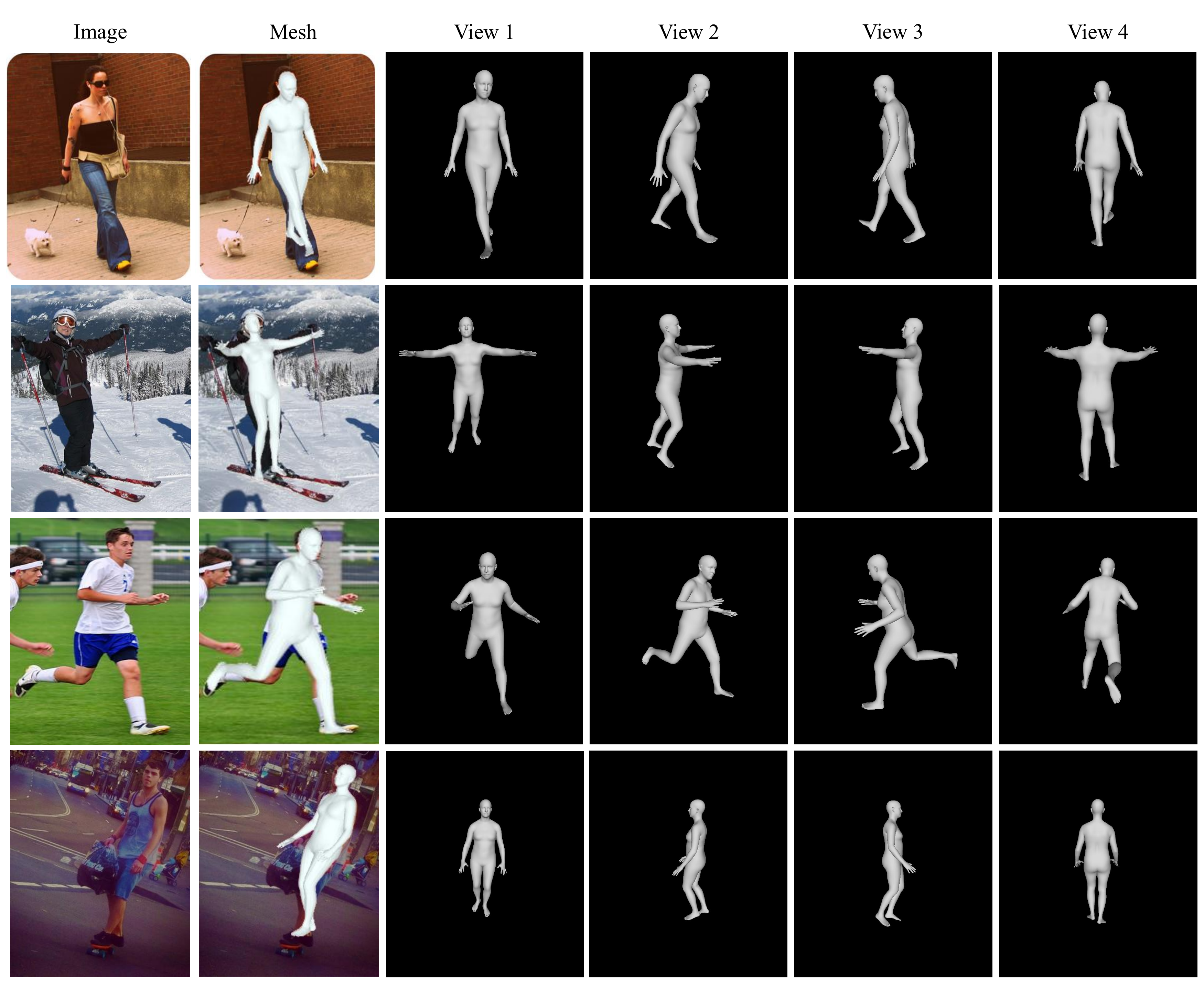}
  \vspace{-10pt}
  \caption{Qualitative results of the proposed GTRS. Images are taken from the in-the-wild COCO ~\cite{lin2014mscoco} dataset.}
  \label{fig:vis}
  \vspace{-5pt}
\end{figure*}

\textbf{Qualitative Results:}
Fig.~\ref{fig:vis} shows the qualitative results of GTRS on in-the-wild images from COCO dataset that GTRS can reconstruct acceptable human meshes. More qualitative results are in \textcolor{blue}{appendix}.

\begin{table}[htp]
\tiny
\centering
  \caption{ Ablation study on different 2D pose input.}
  \resizebox{\linewidth}{!}{
  \begin{tabular}{c|ccc}
\hline
Methods               & Input                         & MPJPE$\downarrow$ & PA-MPJPE$\downarrow$ \\ \hline
Pose2Mesh~\cite{Choi_2020_ECCV_Pose2Mesh}             & GT 2D pose                    & 51.3  & 34.9     \\ \hline
\multirow{4}{*}{GTRS (Ours)} & GT 2D pose                    & 50.3  & 30.4     \\
                      & Estimated 2D pose by ~\cite{get2dpose}           & 64.3  & 47.5     \\ 
                      & Estimated 2D pose by LiteHRNet~\cite{LiteHRNet}   & 66.9  & 48.7     \\\hline
\end{tabular}
}
\label{tab: Ab_noise}
\vspace{-10pt}
\end{table}

\section{Conclusion and Discussion}
We present a lightweight pose-based method, GTRS, for human mesh reconstruction from 2D human pose that reduces Params and FLOPs significantly. A PAM is introduced to exploit structured and implicit joint correlations by using paralleled graph transformers blocks. Then, a MRM is able to combine the extracted pose feature with the mesh template efficiently to reconstruct the human mesh.

Despite GTRS achieving competitive performance, as a pose-based approach, GTRS may not be able to recover varied human body shapes using only 2D human poses as input. Although image-based methods have the potential to reconstruct a more accurate human mesh, pose-based methods are still worth investigating due to their flexibility and lightweight design. In the future, we intend to include another branch that extracts human shape features from images to improve reconstruction capability while keeping the model structure lightweight.

\noindent \textbf{Acknowledgement.} This work is supported by the National Science Foundation under Grant No. 1910844

\newpage
\appendix
\section{Appendix}

\begin{figure*}[!htbp]
\vspace{-10pt}
  \centering
  \includegraphics[width=1\linewidth]{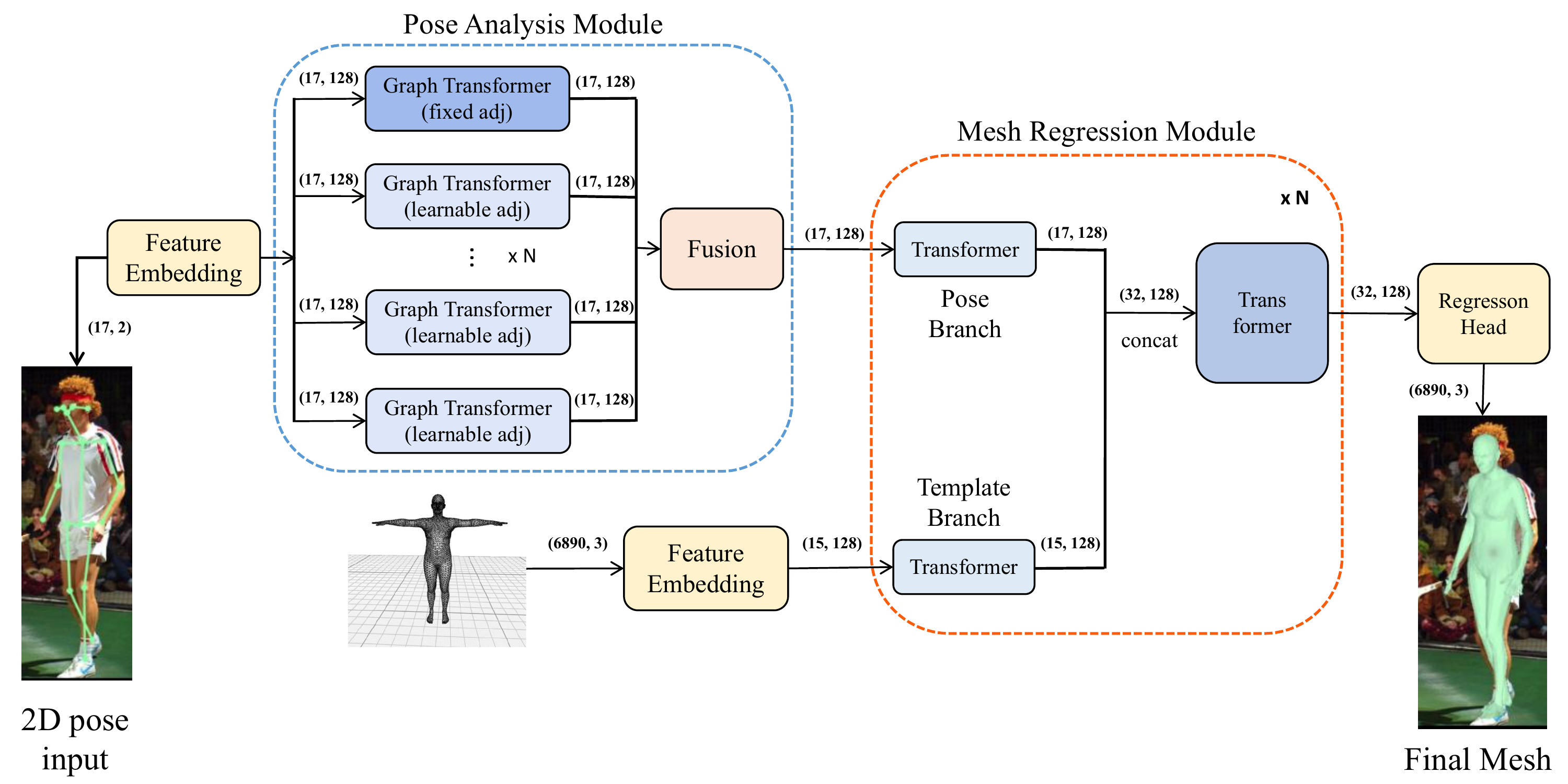}
  \vspace{-15pt}
  \caption{Overview of the proposed GTRS architecture. The mesh template figure is from \cite{lin2021metro}. }
  \label{fig:architectureall}
  \vspace{-5pt}
\end{figure*}

\subsection{Detailed overview of the proposed GTRS architecture}
The detailed overview of GTRS is illustrated in Fig.~\ref{fig:architectureall}. Given the input image, 2D human pose $X_{in\_2D} \in \mathbb{R} ^{17 \times 2} $ is first detected by an off-the-shelf 2D pose detector, then a feature embedding layer (an MLP layer) embeds input 2D pose to  $X_{pose} \in \mathbb{R} ^{17 \times 128}$.

In the pose analysis module, the pose feature $X_{pose} \in \mathbb{R} ^{17 \times 128}$ goes through each parallel graph transformer block without changing the size. A fusion block, consists of convolutional layers, is designed to aggregate these pose features back to  $X^{'}_{pose} \in \mathbb{R} ^{17 \times 128}$. 

Before mesh regression module, the mesh template $ M_{temp} \in \mathbb{R} ^{6890 \times 3} $ is embedded to the template feature  $X_{temp} \in \mathbb{R} ^{15 \times 128}$. In the mesh regression module, the pose feature $X^{'}_{pose}$ and template feature  $X_{temp}$ are modeled separately, then concatenated to the mesh feature $X_{mesh} \in \mathbb{R} ^{32 \times 128}$ for a large transformer modeling. The output would be $X_{out} \in \mathbb{R} ^{32 \times 128} $.  

Finally, the estimated mesh parameter $Y \in \mathbb{R} ^{6890 \times 3} $ can be obtained by the regression head (an MLP layer) for human mesh reconstruction.

\subsection{Design options of transformer block}

\begin{figure}[htp]
\vspace{-5pt}
  \centering
  \includegraphics[width=0.95\linewidth]{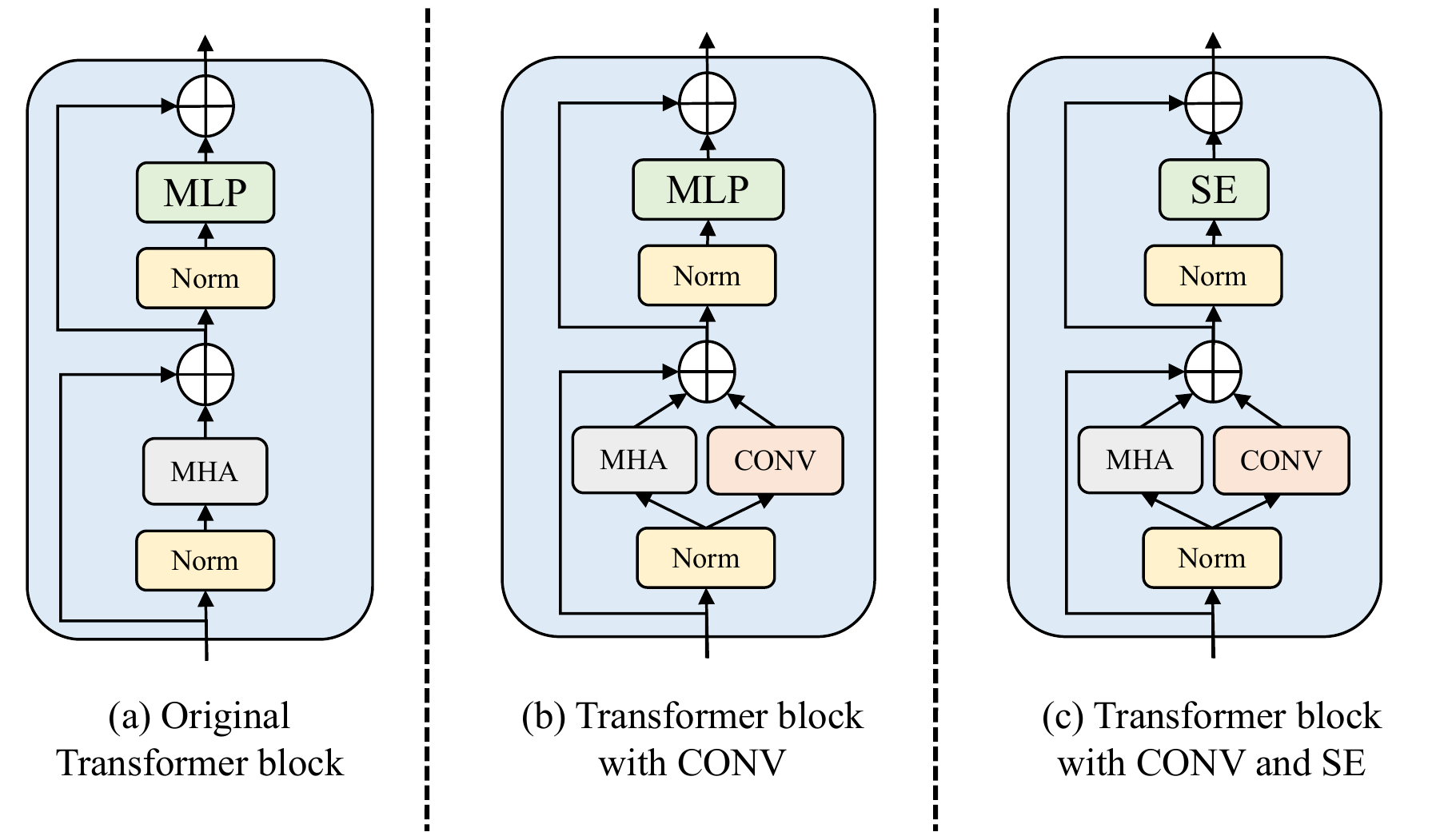}
  \vspace{-5pt}
  \caption{Three design choices for building our proposed transformer block. (a) is the original transformer block from \cite{Dosovitskiy2020ViT}. (b) is the transformer block with adding a convolutional branch parallel to the MHA branch. (c) is the transformer block that replaces MLP by SE block based on (b).}
  \label{fig:transformer}
\end{figure}

\begin{table}[htp]
\tiny
\centering
  \caption{Ablation study on different design options of transformer block. Results are evaluated on Human3.6M dataset.}
  \resizebox{\linewidth}{!}{
  \begin{tabular}{{l|c|c}}
\hline
                               & MPJPE$\downarrow$          & PA-MPJPE$\downarrow$       \\ \hline
Original                       & 65.2          & 47.7          \\ \hline
Add CONV                       & 64.5         & \textbf{47.5}         \\ \hline
Add CONV and replace MLP by SE & \textbf{64.3} & \textbf{47.5} \\ \hline
\end{tabular}
}
\label{tab: transformer_option}
\vspace{-5pt}
\end{table}

Our transformer encoder layer is different from the original transformer encoder in \cite{Dosovitskiy2020ViT}. The structure of our transformer encoder is illustrated in Fig.~\ref{fig:transformer} (c). If we use the original transformer block in Fig.~\ref{fig:transformer} (a), the MPJPE is 65.2 and PA-MPJPE is 47.7 as shown in Table \ref{tab: transformer_option}. Compared with the original transformer blocks, we add one convolutional branch parallel to the MHA branch as shown in Fig.~\ref{fig:transformer} (b). We use this pointwise convolution to create linear combinations of the joint channels while maintaining a low computational cost. The performance has improved since MPJPE is decreased to 64.5 and PA-MPJPE is decreased to 47.5. Then, we replace the MLP with a lightweight SE block \cite{hu2018senet} as shown in Fig.~\ref{fig:transformer} (c) and achieve the best performance (MPJPE is 64.3 and PA-MPJPE is 47.5).


\subsection{Different locations to inject GCN. }

\begin{figure}[htp]
\vspace{-5pt}
  \centering
  \includegraphics[width=0.95\linewidth]{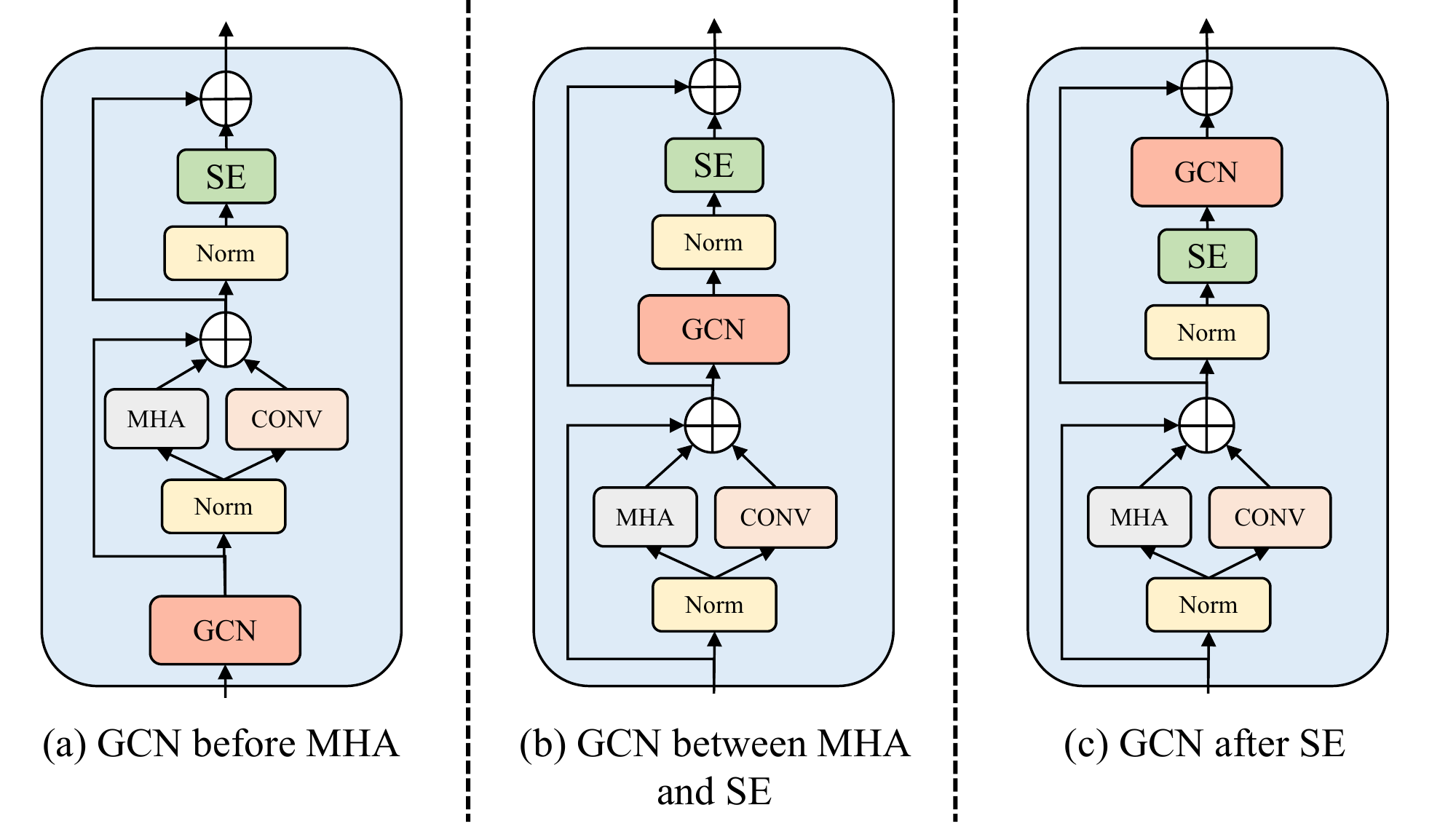}
  \vspace{-5pt}
  \caption{Three different locations to inject GCN to transformer blocks. (a) GCN is in front of the MHA block. (b) GCN is between the MHA and SE block. (c) GCN is behind the SE block. Results are evaluated on Human3.6M dataset. }
  \label{fig:gcn_place}
  \vspace{-5pt}
\end{figure}

In GTRS, we utilize GCNs to maintain a strong joint relationship based on human kinematic information. We inject structural priors provided by GCNs before performing the transformer's multi-head self-attention (MSA) as shown in Fig.~\ref{fig:gcn_place} (a). We also investigate other locations to inject GCNs. The results are reported in Table \ref{tab: gcn_place}. When GCN is between the MSA and SE block as illustrated in Fig.~\ref{fig:gcn_place} (b), the MPJPE is 64.9 and PA-MPJPE is 48.2. When GCN is behind the SE block in Fig.~\ref{fig:gcn_place} (c), the MPJPE is 65.2 and PA-MPJPE is 48.3. We observe that GCN before the MHA achieves the best performance (MPJPE is 64.3 and PA-MPJPE is 47.5), indicating that GCNs do provide structural priors to help final mesh regression. 

\begin{table}[htp]
\tiny
\centering
  \caption{ Ablation study on different locations to inject GCN. Results are evaluated on Human3.6M dataset.}
  \resizebox{0.9\linewidth}{!}{
  \begin{tabular}{l|c|c}
\hline
                       & MPJPE$\downarrow$         & PA-MPJPE$\downarrow$      \\ \hline
GCN before MHA         & \textbf{64.3} & \textbf{47.5} \\ \hline
GCN between MHA and SE & 64.9          & 48.2          \\ \hline
GCN after SE           & 65.2 & 48.3          \\ \hline
\end{tabular}
}
\label{tab: gcn_place}
\vspace{-15pt}
\end{table}

\subsection{Impact of different backbones}
As a pose-based method, GTRS can easily choose a lightweight 2D pose detector to reduce the computational burden. But for most image/video-based methods, a very large feature extractor is needed for extracting features (e.g. ResNet-50 is used by ~\cite{TCMR_Choi_2021,kocabas2020vibe,Kolotouros2019SPIN}). We retrain the I2LMeshNet ~\cite{Moon_I2L_MeshNet} with a small backbone (ResNet18), the results are shown in Table \ref{tab: backbone}. Although selecting a small backbone can reduce the total Params and FLOPs, the performance also drops. GTRS with DARK ~\cite{darkpose_2020} is much more time and resource efficient compared to I2LMeshNet.

\begin{table}[htp]
\tiny
\centering
  \caption{ Impact of different backbones. Results are evaluated on 3DPW dataset.}
  \resizebox{\linewidth}{!}{
\begin{tabular}{c|ccccc}
\hline
                                                               & Backbone & total Params(M) & FLOPs(G) & MPJPE & PA-MPJPE \\ \hline
I2LMeshNet                                                     & ResNet50 & 140.5           & 73.2     & 93.2  & 57.7     \\ \hline
\begin{tabular}[c]{@{}c@{}}I2LMeshNet\\ (small)\end{tabular}   & ResNet18 & 94.6            & 68.3     & 101.8 & 62.9     \\ \hline
\begin{tabular}[c]{@{}c@{}}GTRS with\\ DARK~\cite{darkpose_2020}\end{tabular}       &          & 71.5            & 3.8      & 88.5  & 58.9     \\ \hline
\end{tabular}
}
\label{tab: backbone}
\vspace{-5pt}
\end{table}

\begin{figure*}[htp]
\vspace{-5pt}
  \centering
  \includegraphics[width=0.95\linewidth]{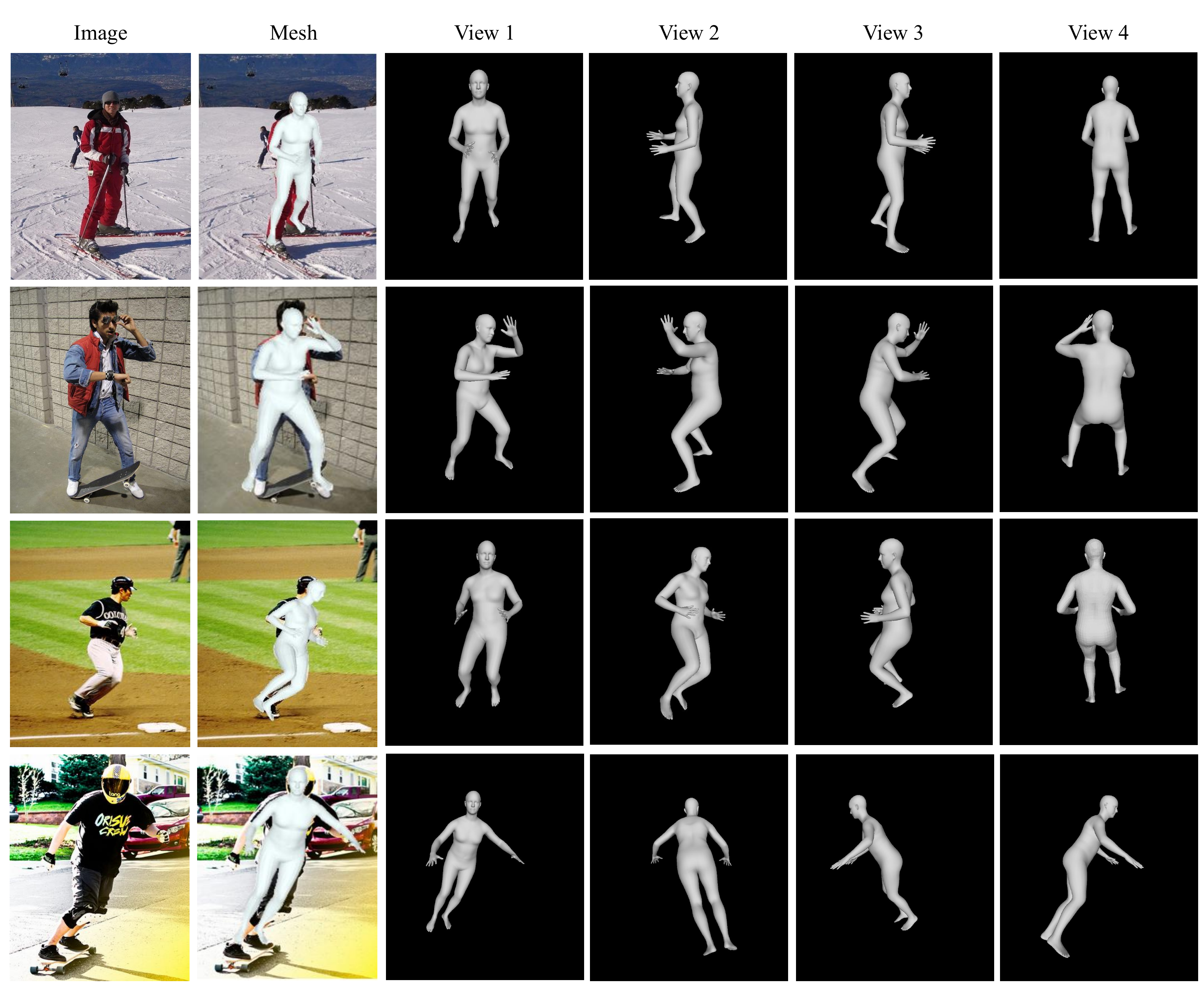}
  \vspace{-5pt}
  \caption{Qualitative results of the proposed GTRS on in-the-wild images. Images are taken from MSCOCO\cite{lin2014mscoco}. }
  \label{fig:vis_all}
  \vspace{-5pt}
\end{figure*}

\subsection{Impact of different losses}
We apply multiple loss introduced in Section 3.6 when training GTRS. Here we evaluate the impact of different losses combination in Table. 

\begin{table}[htp]
\small
\centering
  \caption{ Comparison of the inference speed. The frame per second (fps) is obtained by using batch size 1 on a single GPU/CPU.}
  \resizebox{\linewidth}{!}{
  \begin{tabular}{c|c|c}
\hline
                                                      & Human3.6M & Human3.6M \\ \hline
                                                      & MPJPE     & MPVE      \\ \hline
Vertex Loss only                                      & 71.9      & 86.3      \\ \hline
Vertex Loss + 3D joint Loss                           & 65.3      & 83.4      \\ \hline
Vertex Loss + 3D joint Loss + Normal Loss             & 64.9      & 82.4      \\ \hline
Vertex Loss + 3D joint Loss + Normal Loss + Edge Loss & 64.3      & 82.0      \\ \hline
\end{tabular}
}
\label{tab: fps}
\vspace{-5pt}
\end{table}

Without 3D joint supervision, the MPJPE is 71.9 and MPVE is 86.3 (2D pose is detected by [42]). After adding 3D joint supervision, the performances are boosted to 65.3 of MPJPE and 83.4 of MPVE. The normal loss and edge loss can further improve the performance slightly as shown in the table below.

\subsection{Evaluation on the inference speed}

\begin{table}[htp]
\small
\centering
  \caption{ Comparison of the inference speed. The frame per second (fps) is obtained by using batch size 1 on a single GPU/CPU.}
  \resizebox{\linewidth}{!}{
  \begin{tabular}{l|ccc|ccc}
\hline
      & \multicolumn{3}{c|}{FPS on GPU}                                                                                                               & \multicolumn{3}{c}{FPS on CPU}                                                                                                                \\ \hline
      & \begin{tabular}[c]{@{}c@{}}2D pose \\ detection\end{tabular} & \begin{tabular}[c]{@{}c@{}}3D mesh \\ regression\end{tabular} & overall        & \begin{tabular}[c]{@{}c@{}}2D pose \\ detection\end{tabular} & \begin{tabular}[c]{@{}c@{}}3D mesh \\ regression\end{tabular} & overall        \\ \hline
METRO~\cite{lin2021metro} & -                                                            & -                                                             & 15.65          & -                                                            & -                                                             & 1.81           \\ \hline
GTRS  & 133.73                                                       & 32.41                                                          & \textbf{26.17} & 19.87                                                        & 24.11                                                         & \textbf{10.86} \\ \hline
\end{tabular}
}
\label{tab: fps}
\end{table}

We compare the inference speed between GTRS (end-to-end) and the state-of-the-art image-based mesh reconstruction method METRO\cite{lin2021metro}. The frame per second (fps) is reported in Table \ref{tab: gcn_place}. We use a single NVIDIA RTX 3090 GPU and an AMD Ryzen 3970X 32-Core Processor CPU for testing. Our proposed GTRS is a pose-based method, which means any off-the-shelf 2D pose detector can be easily adopted. Here we use the lightweight OpenPose \cite{osokin2018lightweight_openpose} as the 2D pose detector. The fps numbers of lightweight OpenPose on the testing GPU and CPU are 133.73 and 19.87, respectively. For GTRS, the fps numbers on the GPU and CPU are 32.41 and 24.11, respectively. Overall, GTRS can achieve 26.27 fps on GPU and 10.86 fps on CPU, which is \textbf{much faster} than METRO (the fps is 15.65 on GPU and 1.81 on CPU) on the same computing hardware. Compared to METRO, GTRS gains more advantages on resource-constrained devices since it is significantly more computational efficient.

\begin{figure}[htp]
\vspace{-5pt}
  \centering
  \includegraphics[width=0.9\linewidth]{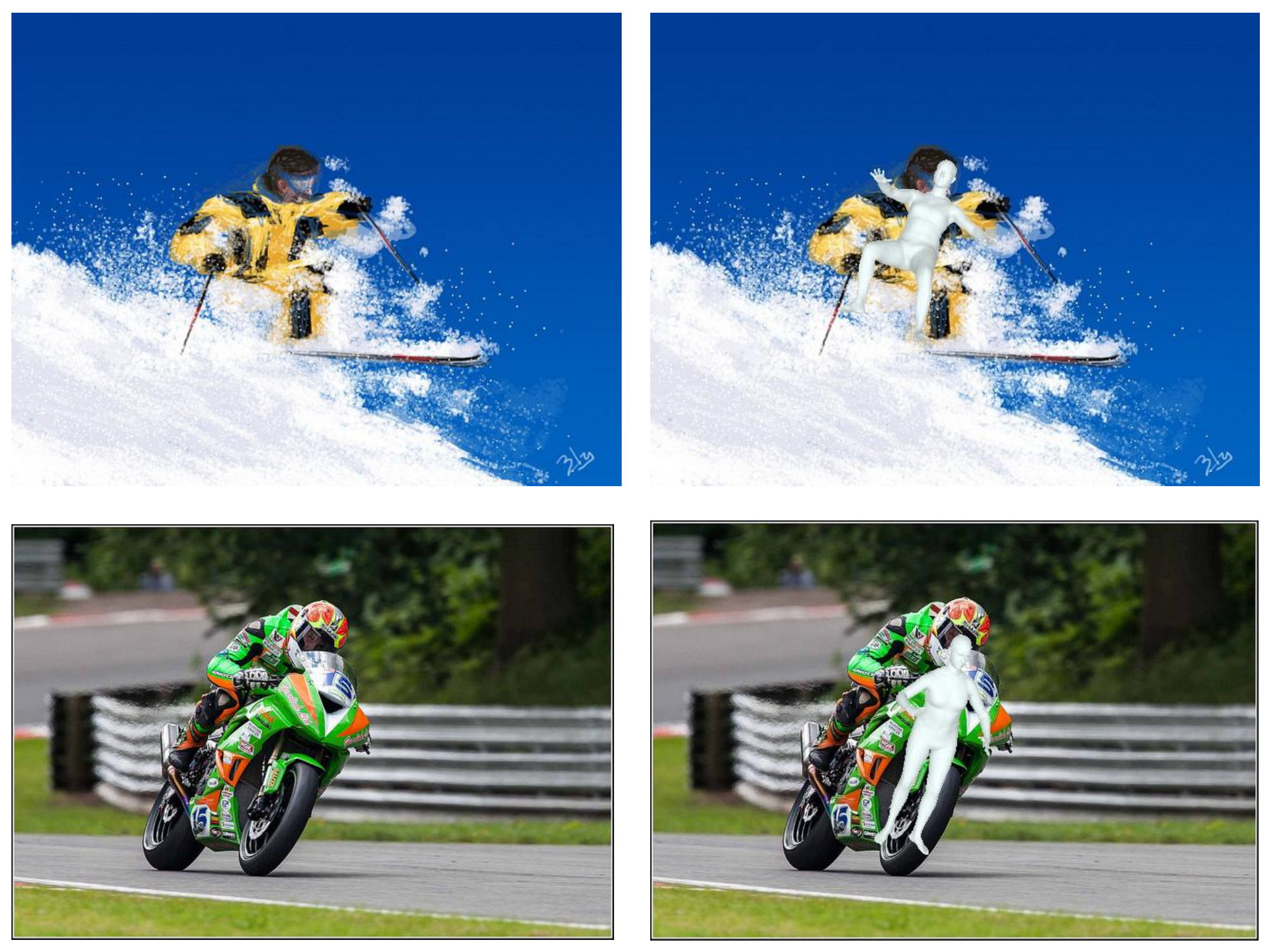}
  \vspace{-5pt}
  \caption{Failure cases due to challenging pose and heavy occlusion. }
  \label{fig:failure}
\end{figure}

\subsection{More qualitative results of GTRS}
In Fig.~\ref{fig:vis_all}, we show the qualitative results of GTRS on in-the-wild images. We observe that GTRS achieves acceptable performance by reconstructing reasonable human mesh on these challenging in-the-wild cases. 
However, there are still some failure cases of our method due to challenging pose and heavy occlusion as shown in Fig.~\ref{fig:failure}.  

\begin{figure}[htp]
\vspace{-5pt}
  \centering
  \includegraphics[width=0.9\linewidth]{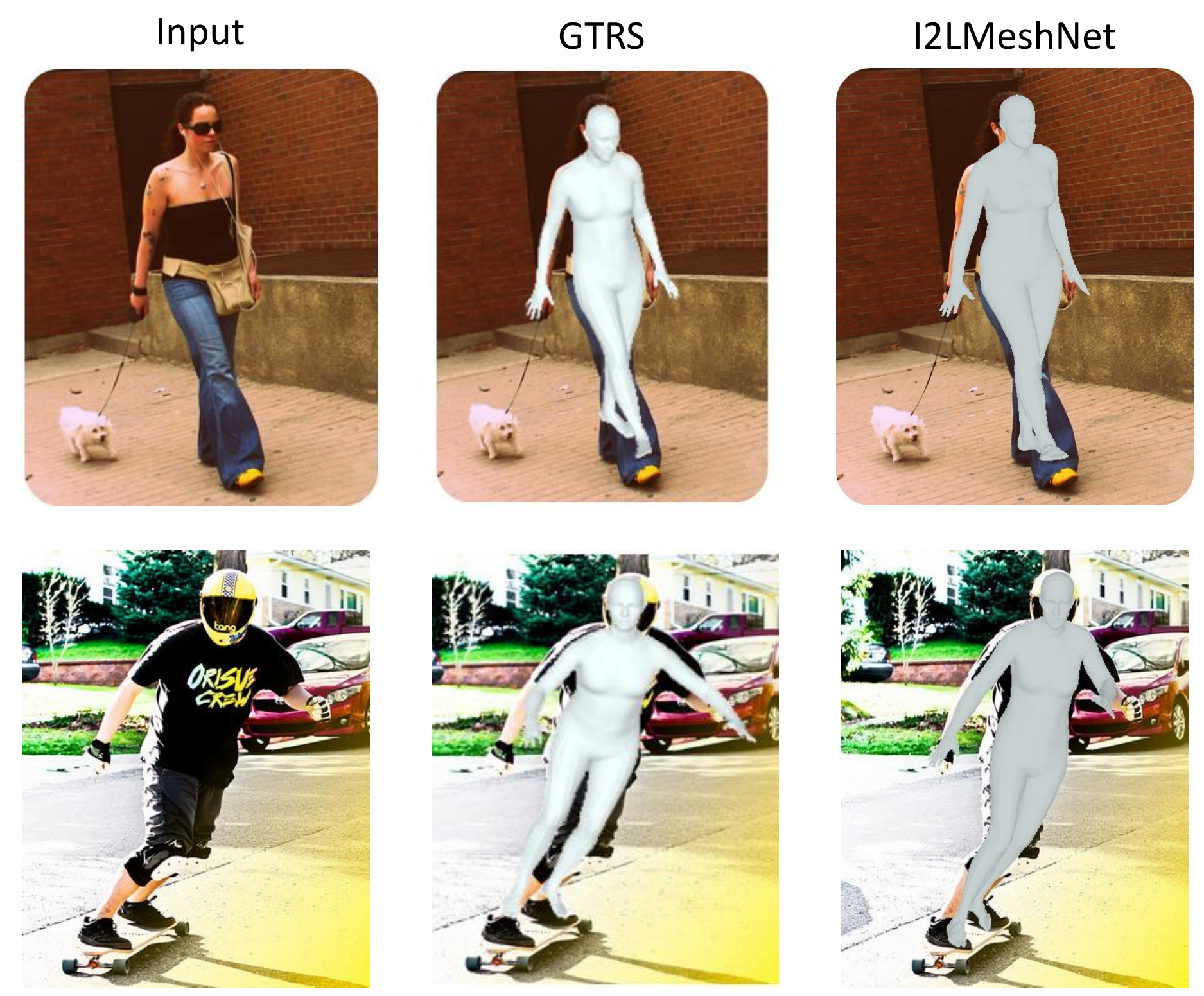}
  \vspace{-5pt}
  \caption{Qualitative comparison with image-based method I2LMeshNet.}
  \label{fig:compare}
\end{figure}

We also compare with image-based method I2LMeshNet \cite{Moon_I2L_MeshNet} in Fig.~\ref{fig:compare}. Our GTRS is comparable with I2LMeshNet while reducing memory and computational cost significantly (only 6.9 \% Params and 1.2 \% FLOPs). 

\newpage

\newpage

\bibliographystyle{ACM-Reference-Format}
\bibliography{egbib}

\end{document}